\documentclass{article}
\usepackage[utf8]{inputenc}
\usepackage{algorithmicx}
\usepackage{graphicx}
\usepackage{algorithm}
\usepackage{algpseudocode}
\usepackage{geometry}
\usepackage{sectsty}
\usepackage[T1]{fontenc}
\usepackage{listings}
\usepackage{caption}
\usepackage{url}
\usepackage{amsmath}
\usepackage[shortlabels]{enumitem} 
\usepackage[english]{babel}
\usepackage{ragged2e}
\usepackage{amsmath}
\usepackage{amssymb}
\usepackage{makeidx}
\usepackage{setspace}
\usepackage{ragged2e}
\usepackage{times}

\usepackage{blindtext}
\usepackage{setspace}
\usepackage{calc}
\usepackage{tocloft}

\usepackage{linegoal}

\makeatletter
\algnewcommand{\LineComment}[1]{\Statex \hskip\ALG@thistlm \(\triangleright\) #1}
\algnewcommand{\LineCommentCont}[1]{\Statex \hskip\ALG@thistlm \parbox[t]{\linegoal}{\hangindent=1em\hangafter=1 $\triangleright$ #1}}

\usepackage[none]{hyphenat}
\usepackage{afterpage}

\sectionfont{\fontsize{20}{20}\selectfont}
\subsectionfont{\fontsize{17}{17}\selectfont}
\subsubsectionfont{\fontsize{15}{15}\selectfont}
 
 \makeatletter
\def\@seccntformat#1{%
  \expandafter\ifx\csname c@#1\endcsname\c@section\else
  \csname the#1\endcsname\quad
  \fi}
\makeatother
\makeindex

\usepackage{tabularx}
    \newcolumntype{L}{>{\raggedright\arraybackslash}X}

\usepackage{caption}
\usepackage{subcaption}

\usepackage{chngcntr}
\counterwithin{figure}{section}
\counterwithin{table}{section}

\begin{document}

\pagenumbering{roman}
\begin{sloppypar}
   \begin{center}
   \begin{large}
       \vspace*{1cm}

       \begin{Huge} {\textbf{B.TECH PROJECT REPORT}} \end{Huge}

       \vspace{0.6cm}
        \begin{Large} on  \end{Large}
        
        \vspace{0.5cm}
        \begin{Huge}{\textbf{Video Anomaly Detection using GAN}} \end{Huge}
        
        \vspace{0.5cm}
        \begin{Large} \textit{By} \end{Large}
        
        \vspace{0.5cm}
        \begin{LARGE}
        \textbf{Anikeit Sethi, 190001003}
        
        \vspace{0.05cm}
        \textbf{Krishanu Saini, 190001029} 

        \vspace{0.30cm}
        \textbf{Sai Mounika Mididoddi, 190001036}
        \end{LARGE}
        
        \vspace{1.2cm}
        \includegraphics[width=.35\textwidth,height=.35\textheight,keepaspectratio]{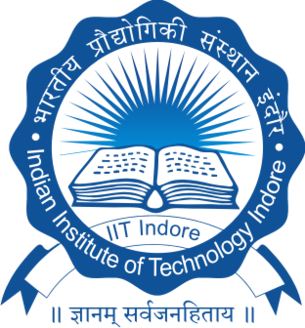}
        
        \begin{Large}
        \vspace{1.0cm}
        \mbox{ \textbf{DISCIPLINE OF COMPUTER SCIENCE AND ENGINEERING,}}
        
        \vspace{0.3cm}
        \textbf{INDIAN INSTITUTE OF TECHNOLOGY INDORE} 
        \end{Large}
        
        \vspace{1.2cm}
        \begin{Large} \textbf{ December 2022} \end{Large}
        
        \newpage
        \thispagestyle{empty}
        \mbox{}
        \newpage

        \begin{LARGE}{\textbf{Video Anomaly Detection using GAN\\}}
        \end{LARGE}
        
        \vspace{1.0cm}
        \begin{Large}{\textbf{A PROJECT REPORT}} \end{Large}
        
        \vspace{0.5cm}
        \textit{Submitted in partial fulfillment of the requirements for the award of the degree of}
        
        \vspace{0.5cm}
        \begin{Large}{\textbf{BACHELOR OF TECHNOLOGY}}\end{Large}
        
        \vspace{0.5cm}
        \textit{in}
        
        \vspace{0.5cm}
        \begin{Large}{\textbf{COMPUTER SCIENCE AND ENGINEERING}}\end{Large}
        
        \vspace{0.5cm}
        \textit{Submitted by:}
        
        \vspace{0.5cm}
        \begin{Large}{\textbf{Anikeit Sethi, 190001003 \\
        \vspace{0.2cm} Krishanu Saini, 190001029 \\
        \vspace{0.3cm} Sai Mounika Mididoddi, 190001036}}\end{Large}
        
        \vspace{0.7cm}
        \begin{Large}{\textbf{Discipline of Computer Science and Engineering, \\ \vspace{0.2cm}
        Indian Institute of Technology Indore}}\end{Large}
        
        \vspace{0.7cm}
        \textit{Guided by:}
        
        \vspace{0.4cm}
        \begin{Large}{\textbf{Dr. Aruna Tiwari, \\ \vspace{0.2cm}
            Professor, \\ \vspace{0.2cm}
            Computer Science and Engineering, \\ \vspace{0.2cm}
            IIT Indore}}\end{Large}
        
        \vspace{0.3cm}
        \includegraphics[width=2cm,height=3cm,keepaspectratio]{images/Logo.png}
        
        \vspace{0.5cm}
        \begin{Large}{\textbf{INDIAN INSTITUTE OF TECHNOLOGY INDORE \\ \vspace{0.4cm}
         December 2022}}\end{Large}
    \end{large}        
    \end{center}
    
    \newpage
    \thispagestyle{empty}
    \mbox{}
    \newpage
    
    \begin{center}  \begin{Large}
    \addcontentsline{toc}{section}{Candidate's Declaration}
        \section*{\textbf{Candidate's Declaration}}
    \end{Large}  \end{center}
    \vspace{0.5cm}
    
    \begin{spacing}{1.35}
    \begin{large}
    We hereby declare that the project entitled \textbf{``Video Anomaly Detection using GAN"} submitted in partial fulfillment for the award of the degree of Bachelor of Technology in Computer Science and Engineering completed under the supervision of \textbf{Dr. Aruna Tiwari, Professor, Computer Science and Engineering,} IIT Indore is an authentic work. \\ \\ \qquad Further, we declare that we have not submitted this work for the award of any other degree elsewhere.  \\ \\ \\ 
    \begin{flushright}
    Anikeit Sethi  \\ Krishanu Saini \\   Sai Mounika Mididoddi \\
    \end{flushright}
    \end{large}
    
    \rule{\textwidth}{0.01cm}
    
    \begin{center}
    \begin{Large}
    \addcontentsline{toc}{section}{Certificate by BTP Guide}
        \section*{\textbf{Certificate by BTP Guide}}
    \end{Large}
    \end{center}
    \vspace{0.5cm}
    \begin{large}
    It is certified that the above statement made by the students is correct to the best of my/our knowledge.\\ \\ \\ \\
    
    \begin{flushright}
    Dr. Aruna Tiwari \\
    Professor \\
    Discipline of Computer Science and Engineering \\
    IIT Indore \\
    \end{flushright}
    \end{large}
    \end{spacing}
    
    \newpage
    \thispagestyle{empty}
    \mbox{}
    \newpage
    
    \newpage
    \begin{center}  \begin{Large}
    \addcontentsline{toc}{section}{Preface}
        \section*{\textbf{Preface}}
    \end{Large}  \end{center}
    \vspace{0.5cm}
    \begin{Large} 
    \begin{spacing}{1.35}
    This report on \textbf{``Video Anomaly Detection using GAN"} is prepared under the guidance of Dr. Aruna Tiwari, Professor, Computer Science and Engineering, IIT Indore. \\ \\
    Through this report, We have tried to provide a detailed description of our approach, design, and implementation of the method to address
the problem of Anomaly Detection using Generative adversarial networks (GANs). We have processed the videos into frames and use them for detecting anomalies. We explained the proposed solution to the best of our abilities.\\ \\ \\
    \textbf{Anikeit Sethi \\
    Krishanu Saini \\
    Sai Mounika Mididoddi}\\ 
    B.Tech. IV Year \\
    Discipline of Computer Science Engineering \\
    IIT Indore \\
    \end{spacing}
    \end{Large}
    
    \newpage
    \thispagestyle{empty}
    \mbox{}
    \newpage
    
    \newpage
    \begin{center}  \begin{Large}
    \addcontentsline{toc}{section}{Acknowledgments}
        \section*{\textbf{Acknowledgments}}
    \end{Large}  \end{center}
    \vspace{0.5cm}
    \begin{Large} 
    \begin{spacing}{1.35}
    We want to thank our B.Tech Project supervisor, Dr. Aruna Tiwari, for her guidance and constant support in structuring the project and for providing valuable feedback throughout this project. Furthermore for giving us an opportunity to deal with advanced technologies. We would also like to thank all the faculty members of the Discipline of Computer Science and Engineering for their invaluable support during the presentations. We are grateful to the Institute for providing the necessary tools and utilities to complete the project. We wish to express our sincere gratitude to Mr. Rituraj for his guidance throughout the project and for helping us at every stage. Lastly, we offer our sincere thanks to everyone who helped us complete this project, whose name we might have forgotten to mention.\\ \\ \\
    \textbf{Anikeit Sethi \\
    Krishanu Saini \\
    Sai Mounika Mididoddi}\\ 
    B.Tech. IV Year \\
    Discipline of Computer Science Engineering \\
    IIT Indore \\
    
    \newpage\thispagestyle{empty}
    \mbox{}
    \newpage
    \end{spacing}
    \end{Large}

\begin{center}
    \begin{LARGE}
    \addcontentsline{toc}{section}{Abstract}
        \section*{Abstract}
    \end{LARGE}
\end{center}
\begin{Large}
\begin{spacing}{1.25}
     Accounting for the increased concern for public safety, automatic abnormal event detection and recognition in a surveillance scene is crucial. It is a current open study subject because of its intricacy and utility. The identification of aberrant events automatically, it's a difficult undertaking because everyone's idea of abnormality is different. A typical occurrence in one circumstance could be seen as aberrant in another. Automatic anomaly identification becomes particularly challenging in the surveillance footage with a large crowd due to congestion and high occlusion. With the use of machine learning techniques, this thesis study aims to offer the solution for this use case so that human resources won't be required to keep an eye out for any unusual activity in the surveillance system records.\\
\begin{spacing}{1.25}We have developed a novel generative adversarial network (GAN) based anomaly detection model. This model is trained such that it learns together about constructing a high-dimensional picture space and determining the latent space from the video's context. The generator uses a residual Autoencoder architecture made up of a multi-stage channel attention-based decoder and a two-stream, deep convolutional encoder that can realise both spatial and temporal data. We have also offered a technique for refining the GAN model that reduces training time while also generalising the model by utilising transfer learning between datasets. Using a variety of assessment measures, we compare our model to the current state-of-the-art techniques on four benchmark datasets. The empirical findings indicate that, in comparison to existing techniques, our network performs favourably on all datasets.
\end{spacing}
\end{spacing}
\end{Large}
\newpage\thispagestyle{empty}
    \mbox{}
    \newpage

\tableofcontents{}
\newpage
\addcontentsline{toc}{section}{List of Figures}
\setlength{\cftparskip}{1.2\baselineskip}
\listoffigures
\newpage\thispagestyle{empty}
\addcontentsline{toc}{section}{List of Tables}
\listoftables
\newpage\thispagestyle{empty}
\pagenumbering{arabic}
\begin{center}
    \begin{LARGE}
        \textbf{Chapter 1}
        \vspace{-0.5cm}
        \section{Introduction}
        \label{Introduction}
    \end{LARGE}
\end{center}

\begin{Large}
\begin{spacing}{1.0}
The need for security is becoming more and more in the modern world, making video surveillance a significant everyday worry. Due to the popularity of this demand, several cameras that record a lot of footage have been installed in many places. The majority of current video surveillance systems are entirely controlled by people. It takes a lot of effort and time to monitor video. Large-scale videos cannot reveal strange events to a person. However, even a minor error might result in an intolerable loss. Therefore, it is crucial to create a system that can handle several video frames and find anomalies in them. Therefore, extensive research is being done on automated video surveillance. Building security, traffic analysis, video monitoring, and other surveillance scenarios are some of the key applications of automated abnormal event detection and recognition. \\ \\
The notion of abnormal is ambiguous or context-dependent, making automatic abnormal event identification difficult. Various scholars have employed methodologies based on supervised and unsupervised learning. Since it is impractical to produce labels for every category of aberrant behaviour, general-purpose abnormality detection via supervised learning may not be feasible. However, it is feasible to develop labels for aberrant occurrences and utilise supervised learning depending on the needs at a certain location. For instance, automobile access is unusual in regions where only pedestrians are permitted. A machine learning-based approach to detect abnormal events can save a lot of time and effort because they happen less frequently than regular occurrences. As a consequence, we provide a method for anomaly identification using a Generative Adversarial Network.
in surveillance footage. We further discuss the motivation for the work
in the subsequent section, followed by the objectives of the work. \\ \\
   
\vspace{0.5cm}     
\subsection{Motivation}
\label{Motivation}
The identification and categorization of aberrant events is still an active research field. It is not possible to identify numerous anomalous occurrences in surveillance footage using a completely autonomous approach. There are a lot of publicly accessible real-world benchmark datasets for study. Closed-circuit television (CCTV) cameras are almost prevalent, and video surveillance is a fundamental requirement when thinking about people's security. The fundamental source of motivation for our work on automatic identification and categorization of anomalous occurrences in surveillance scenes is the production of an enormous volume of surveillance footage and growing concerns about security and safety. 

\subsection{Objectives}
\label{Objectives}
The objective of our B.Tech. project is to implement GAN based method to detect and classify anomalous events in surveillance scenes. The different
sub-objectives to achieve the objective are described below: 
\begin{enumerate}[a)] 
    \item \textit{Preprocessing videos from dataset to be used as an input for designed models:} This involves generating frames from all the videos of the considered dataset.
    \item \textit{Extracting features from the preprocessed data:} The frames extracted are used to generate features using the AlexNet model pre-trained on the ImageNet dataset.
    \item \textit{Develop an Architecture for the Generator:} We first model an
    Encoder-Decoder architecture, which we use as Generator for our
    GAN model. The generator takes an image, encodes it to a latent space, and then decodes it, giving an image of the same size as the input.
    \item \textit{Include discriminator(s) to build the GAN Model:} A GAN model
    has two submodels, Generator and Discriminator. After finalizing
    the Generator Architecture, discriminator(s) are added to complete
    our GAN model.
    \item \textit{Demonstrate the efficiency of our model on different Datasets:} Once the training is done, the model is tested on the UMN dataset, UCSD-Peds
    dataset, Avenue dataset, Subway entrance and exit dataset using various evaluation metrics.
    
\end{enumerate}
In this chapter \ref{Introduction}, we have discussed the problem statement and related things in detail. We will be describing anomaly detection techniques, deep learning models and key aspects in the next chapter \ref{LiterRev}.
\end{spacing}
\end{Large}
\begin{center}
    \begin{LARGE}
        \textbf{Chapter 2}
        \vspace{-0.5cm}
        \section{Literature Review}
        \label{LiterRev}
    \end{LARGE}
\end{center}

\begin{Large}
\begin{spacing}{1.0}
The critical work that has been offered in this field is all represented by the approaches we explain in this chapter for the detection and recognition of abnormal events. Image feature extraction is a fundamental step before any anomaly detection. The selection of key-frames is essential for anomaly identification. For the same, we might make use of transformers, pixel-level differences, and uniform or random sampling of frames. Additionally, we have experimented with object-focused background removal. The principles of Generative Adversarial Networks and their extensions, as well as how they were utilised to discover abnormalities, will also be covered in the sections that follow.  \\

\subsection{Video Preprocessing}
Data preprocessing is a crucial prerequisite for cleaning the data and preparing it for a machine learning model, which also improves training accuracy and efficiency. Certain ways of processing video datasets have been introduced \cite{Gowda2021SMARTFS, PCA1901}. These need turning video footage into a series of frames and then executing further transformations.

\subsubsection{Video Fragmentation}
\label{sec:video_frag_lit}

 In a video, not all the frames are essential, and there are many redundant frames. So, it is important to fragment a video. The ability to provide a thorough and concise key-frame-based summary \cite{Gowda2021SMARTFS} of a video is made possible by the fragmentation of a movie into visually and temporally coherent pieces and the extraction of a sample key-frame for each designated fragment. The created summary using video fragmentation and key-frame extraction is significantly more effective for learning the video content and carrying out anomaly identification than simple methods that sample video frames with a constant step. \\ \\
Various methods of selecting frames are discussed below:
\begin{enumerate}
\item Uniform Sampling: We try to pick frames after uniform intervals. But this method is considered to be less effective as many key-frames might be lost in between the intervals.
\item Random Sampling: We try to choose random frames instead of choosing uniformly. This eliminates any kind of bias since each frame has an equal probability of selection. 
\item Pixel level difference; Another way to choose frames is by taking frames whose pixel level difference is more than a threshold value. This reduces the chances of taking redundant frames and also decreases the chances of missing key-frames by a significant amount.
\item Key-Frame Selection: We can make use of features extracted for choosing only those frames which are important for summarizing a video. we could use transformers which generate segmentation masks for this purpose which are discussed below.
\end{enumerate}   

\subsubsection{Feature Extraction}
\label{Feature Extraction}
The major steps in data pre-processing include data augmentation, frame generation and feature
extraction. In our experiment, we considered the UCSD dataset for this experimentation. The videos are processed to generate frames of dimensions required to facilitate the network needs. Feature extraction is performed on these video frames to identify distinguishing characteristics. We use a pre-trained model of AlexNet to extract features. These models process the input to give an output vector of 1000 values. The following figure \ref{Fig: feature extraction } summarizes the process described above.  \\
\begin{figure}[h]
\centering
\includegraphics[width=.75\textwidth,height=.75\textheight,keepaspectratio]{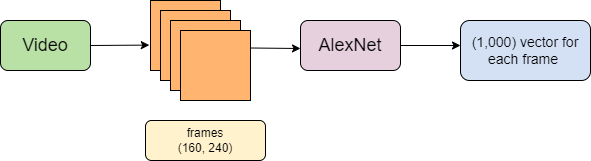}
\captionof{figure}{\large Flow of Feature Extraction}\label{Fig: feature extraction }
\end{figure} 

There are eight learnable levels in Alexnet (figure \ref{Fig: alexnet architecture }). RELU activation is used in each of the five levels of the model, with the exception of the output layer, which uses max pooling followed by three fully connected layers. Prior to choosing the frames, it is crucial that we lower the dimensionality of our data samples.
\begin{figure}[h]
\centering
\includegraphics[width=.75\textwidth,height=.75\textheight,keepaspectratio]{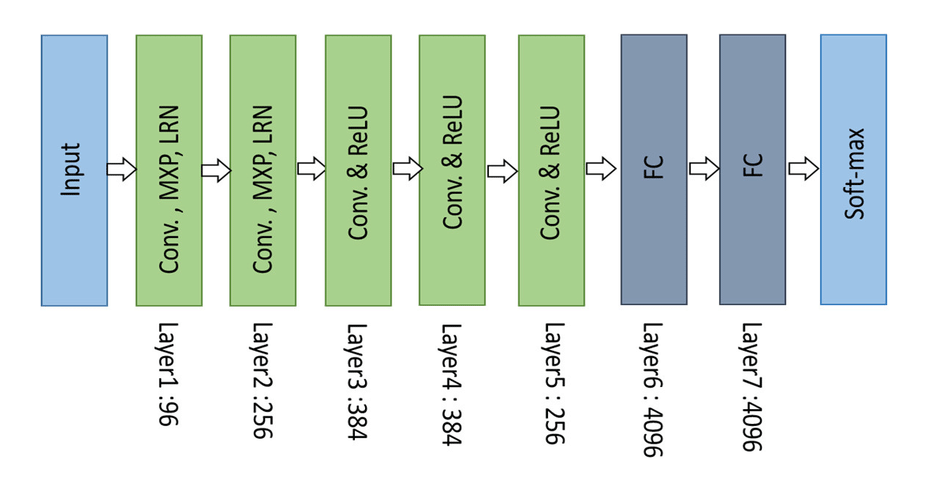}
\captionof{figure}{\large AlexNet architecture}\label{Fig: alexnet architecture }
\end{figure}

\begin{itemize}
    \item \textbf{Transformers} \\
    \label{Transformers}
Transformers \cite{Gowda2021SMARTFS} have recently had a significant influence on several computer vision tasks. Numerous vision jobs are replacing CNNs with transformers as a result of the enormous success of the detectron algorithm, which has introduced a new paradigm to the object detection problem. Many add an additional classification token to the input of transformers for classification tasks in both NLP and computer vision. Since the encoders are mostly made of self-attention, all of the input tokens interact with one another, making it possible to identify the class of the entire input using the classification token. Max-DeepLab combines the transformer and the CNN by making both of their feedback by using a number of memory tokens identical to the previous categorization token. We utilize this concept of memory tokens in the videos. Each frame operates independently while exchanging information with intermediate communications thanks to inter-frame communication transformers. While execution independence across frames speeds up inference, communications increase accuracy.\\

\begin{figure}[h]
\includegraphics[width=\textwidth,height=\textheight,keepaspectratio]{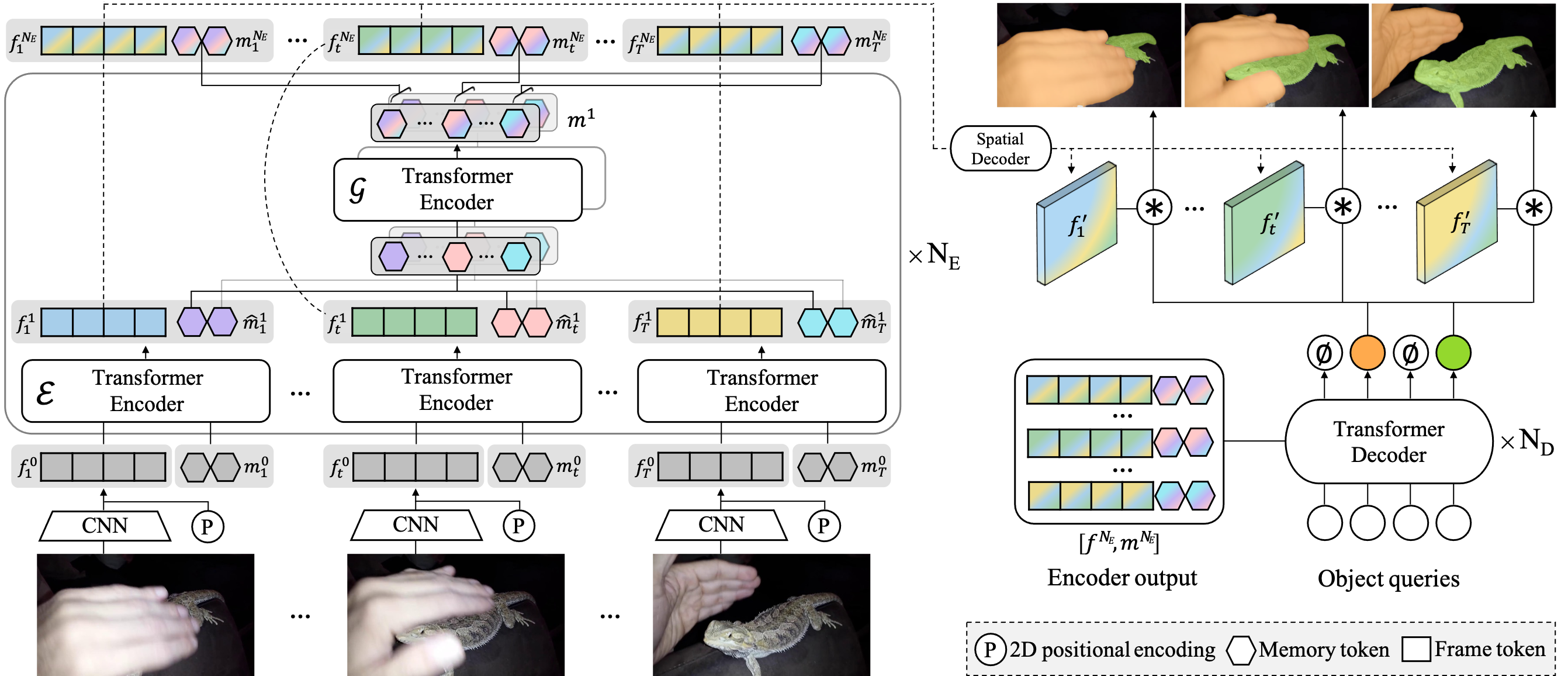}
\captionof{figure}{\large Overview of IFC framework \cite{Gowda2021SMARTFS}}\label{Fig: Overview of IFC framework }
\end{figure}
\end{itemize}

\subsubsection{Feature Reduction}

\label{PCASection}
The difficulty of computing increases with the number of features in machine learning situations. The numerous effects of high-dimensional feature spaces are covered under the "curse of dimensionality." The term was first used by Richard E. Bellman et al.~\cite{curseOfDimensionality1961}. Data points in high dimensional space can be considered to be extremely sparse, i.e., every data point is technically distant from the rest of the points. Distance measures might not be useful in such circumstances. Additionally, it is possible that certain elements will be unimportant and will reduce the impact of informative features. \\ \\ 
By using dimensionality reduction, features may be moved from a high-dimensional space to a low-dimensional space while still keeping crucial data intact. Both cluster analysis and data visualisation are aided by this. Dimensionality reduction techniques include Linear Discriminant Analysis (LDA), Principal Component Analysis (PCA), Generalized Discriminant Analysis (GDA), non-negative matrix factorization, etc. \\ \\ 

\begin{itemize}
    \item \textbf{Principal Component Analysis} \\
    Principal Component Analysis was introduced by Karl Pearson et al.~\cite{PCA1901}. Maximizing variance in low-dimensional space is the method's main goal. The data points are attempted to be modelled in terms of additional variables known as main components. The direction of the maximum amount of variation in the data points is represented by the new variables. Only the variables with the largest variance are picked as the main components out of all such calculated variables. These few variables contain the majority of the data points' information, hence in order to have unique information captures, we strive to choose uncorrelated variables as the main components.\\ \\
Below are the steps involved in PCA ~\cite{PCA1901}: 

\begin{enumerate}
    \item \textit{Standardization}: This step standardizes the feature variables so that all the values contribute equally to the analysis. It transforms all variables to the same scale.
        \begin{table}[h!]
        \large
        \centering
        \begin{tabular}{l l} 
        \hline
        Notation & Description\\ [0.5ex] 
         \hline
          $x_{ij}$ & $i^{th}$ data point value for $j^{th}$ feature\\
          $\mu_j$ & mean of $j^{th}$ feature across all data points\\
          $\sigma_j$ & standard deviation for $j^{th}$ feature across all data points \\ 
          $z_{ij}$ & value of $x_{ij}$ after standardization\\[1ex]
         \hline
        \end{tabular}
        \caption{\large Major Math Symbols - Standardization}
        \end{table}
            \[  z_{ij} = \frac{x_{ij}-\mu_j}{\sigma_j}\]
    
    \item \textit{Covariance Matrix Computation}: Correlation between variables indicates information redundancy. A covariance matrix is a table that contains the covariance values of all possible pairs of features. If the data points have $n$ features, then the covariance matrix is an $n \times n$ matrix. In the covariance matrix, the values on the diagonal of the matrix indicate the variance of individual features.
    \item \textit{Eigenvector and Eigenvalue Computation}: Principal components are chosen to maximize the variance of the data points. The covariance matrix calculated in the previous step is given as an input to calculate eigenvectors. These eigenvectors are in the direction of maximum variance. So they can be chosen as the principal components. The variance along these principal components is given by the value of the eigenvalue associated with each of these eigenvectors. The greater the eigenvalue, the greater the variance, the more significant is the corresponding principal component. Arrange the eigenvectors using the eigenvalues and pick the variables with the highest eigenvalues as the principal components.
    \item \textit{Recast data points along the principal components}: In the last step, the aim is to recast the data points from the original axes along the principal components. The feature vector matrix is constructed using the principal components decided in the previous step.
    
    \[ \text{Final DataSet} = \text{Feature Vector}^T \times \text{Standardized DataSet}\]
\end{enumerate} 

We will now understand the advantages of using PCA for dimensionality reduction in the upcoming section \ref{AdvPCA}.\\

\noindent \textbf{Advantages of Principal Component Analysis}
\label{AdvPCA}
\begin{itemize}
  \item It aids in improved cluster analysis and data visualisation.
  \item It enhances the training and performance of the ML model by removing correlated variables (that don't contribute to any decision-making) and hence redundancy in data.
  \item Principal components are easily computable since it uses linear algebra.
  \item Through a reduction in the number of characteristics, overfitting is avoided.
  \item High variation among the new variables is the consequence, which enhances data presentation and reduces noise.
\end{itemize}
\end{itemize}

\subsection{Video Classification- Deep Learning}
\label{sec:deep_learning_lit}

Most cutting-edge computer vision solutions for diverse tasks are built around convolutional networks. Since 2014, very deep convolutional networks have begun to gain traction and have significantly improved across a range of benchmarks. As long as sufficient labelled data is available for the training, increased model size and computational cost tend to result in immediate quality improvements for most tasks. 
Video understanding poses particular difficulties for machine learning methods. In addition to the spatial features found in 2D images, the video also offers the additional (and intriguing) aspect of temporal features. While this more data gives us more material to work with, it also necessitates new network topologies and frequently increases memory and computing requirements.
However, computational efficiency and low parameter count still enable various use cases. Here, using appropriately factorised convolutions and aggressive regularisation, we can investigate strategies to scale up networks that utilise the additional processing as effectively as possible.\\

\noindent We will be taking a look at a few deep-learning techniques in the next section.
\subsubsection{Early Methods for Video Classification}
Deep learning has enabled the autonomous extraction of features and patterns from datasets, making it possible to extract valuable spatial information hidden in the data. Early video classification methods employed convolutional neural networks, which learn patterns and transform the data. Some of these techniques are listed below: 

\begin{itemize}
    \item{\textbf{Inception Architecture- Classify single frame at a time}  
    
     In order to identify each clip using our first method, we will disregard the temporal characteristics of the video and focus only on a single frame from each clip. CNN used to do this, More specifically, InceptionV3 \cite{szegedy2016rethinking}.
     
     An image classification network is the simplest way to accomplish video classification. Every video frame will now be subjected to an image classification model, and the final probabilities vector will be obtained by averaging all the individual probabilities. This method works very well, and we will use it in this post.
    
    Additionally, it is important to note that movies typically have many frames. As a result, a select few frames dispersed throughout the entire video need to be subjected to a classification model.}

    \item {\textbf{Late Fusion}  
    
    In practice, the Late Fusion\cite{amir2003ibm} method is similar to the Single-Frame CNN method but a little more difficult. The Single-Frame CNN method differs because it calculates an average of all predicted probabilities after the network has finished its job. The Late Fusion approach still incorporates averaging (or another fusion technique) into the network. As a result, the frame sequence's temporal structure is also considered.
    \begin{figure}[h]
        \centering
        \includegraphics[width=1.0\textwidth,height=1.0\textheight,keepaspectratio]{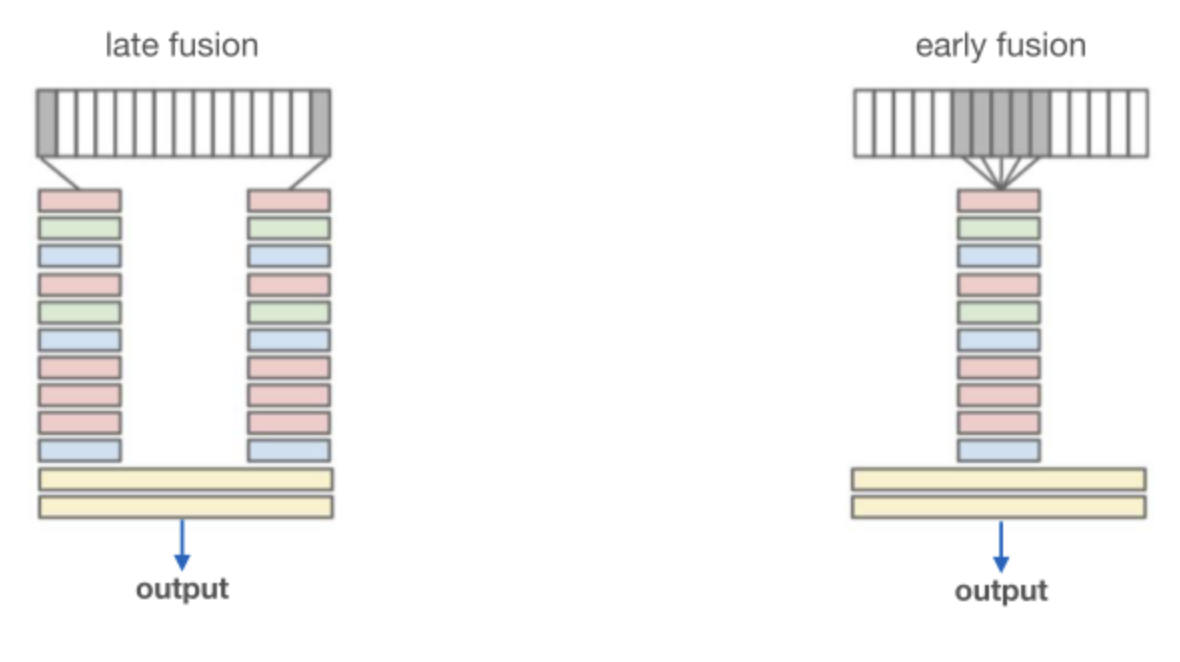}
          \captionof{figure}{\large Early Fusion and Late Fusion}\label{elf} 
      \end{figure}
    The output of many networks that operate on distant temporal frames is combined using a fusion layer. The maximum pooling, average pooling, or flattening techniques are typically used to execute it.
    With this method, the model can acquire spatial and temporal details on the look and motion of the objects in a scene. Each stream independently classifies each image (frame), and then the projected scores are combined using the fusion layer.}

    \item {\textbf{Early Fusion}

    This method differs from late fusion in that the video's temporal and channel (RGB) dimensions are fused at the outset before being passed to the model. This enables the first layer to work over frames and discover local pixel motions between adjacent frames.\cite{amir2003ibm}
    An input video with the dimensions (T x 3 x H x W), three RGB channel dimensions, and after fusion of two spatial dimensions H and W, is transformed into a tensor with the dimensions (3T x H x W).
    \nocite{iyengar2003discriminative}}

\end{itemize}

\subsubsection{Modern Methods}

\begin{itemize}
\item{\textbf{Deep Bi-Directional LSTM - CNN with LSTM}}

\begin{figure}[h]
    \centering
    \includegraphics[width=1.0\textwidth,height=1.0\textheight,keepaspectratio]{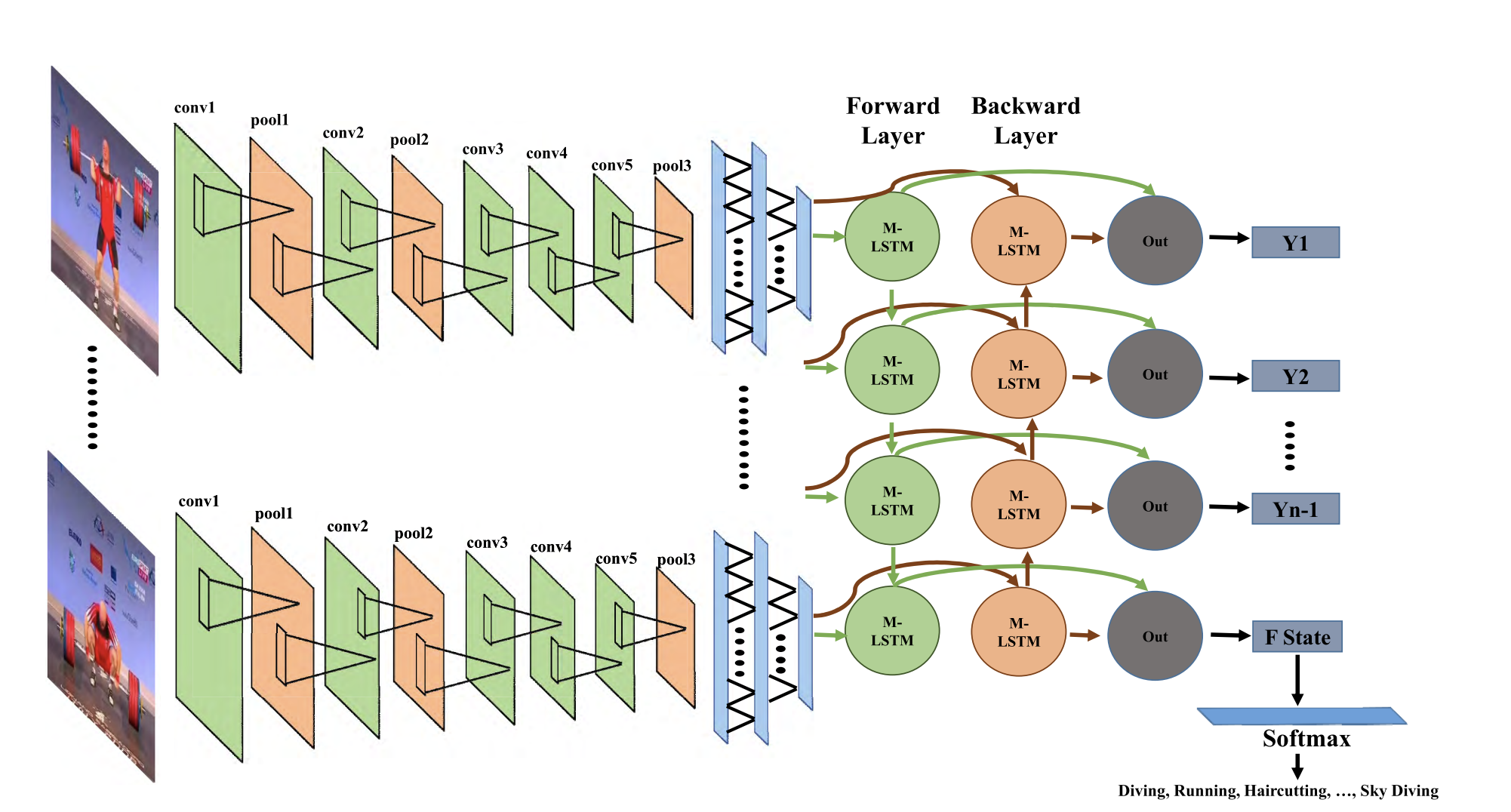}
      \captionof{figure}{\large DB-LSTM \cite{ullah2017action}}\label{Fig:dblstm} 
  \end{figure}
Convolutional networks are used in this method to extract the local features of each frame. To temporarily combine the retrieved information, the outputs of these separate convolutional networks are input into a many-to-one multilayer LSTM \cite{ullah2017action}network. Refer to Figure \ref{Fig:dblstm}

Some approaches \cite{nawaratne2019spatiotemporal,wei2019detecting} used an autoencoder architecture that used both stacked convolutional neural network layers to learn the spatial structure and a stacked convolutional LSTM to learn the temporal representation and the normal events to extract the appearance feature as well as the motion feature from the video input.

\item {\textbf{Channel Attention Module}}
\label{sec:attention_pre}

To understand the working of attention, it is crucial to understand the problem it solved. It was first introduced for problems dealing with time-series tasks.
Before attention, models such as RNN with LSTM and GRU were prevalent. Such frameworks worked very well, especially with LSTM and GRU components. However, as the size of the time series dataset increased, the performance of such models dropped due to the vanishing gradient problem. This shortcoming led to the development of attention. \\ \\

 \begin{figure}[h]
    \centering
    \includegraphics[width=0.5\textwidth,height=0.5\textheight,keepaspectratio]{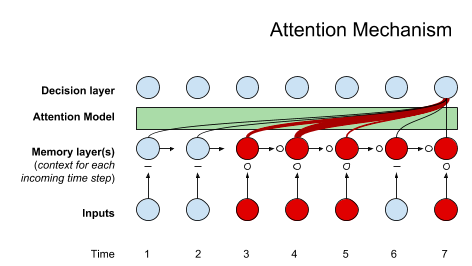}
      \captionof{figure}{\large Attention Module Architecture}\label{Fig:Attention2} 
  \end{figure}

\noindent Attention is derived from the biological model of the human brain—the cognitive ability of our brain to focus on more subtle but important parts of an event. It was introduced for computer vision in Larochelle and Hinton \cite{Larochelle2010LearningTC}. The core idea was to establish a direct connection with each of the images in the sequence. Looking at the distinct frames together, one can infer much more about the video clip than from a single image.

 \item{\textbf{3D CNN}}
 
This method uses a 3D convolution network to handle both temporal and spatial data using a 3D CNN. The Slow Fusion Method is another name for this process. This approach slowly merges temporal and spatial data at each CNN layer over the whole network, in contrast to early and late fusion.
\begin{figure}[h]
    \centering
    \includegraphics[width=0.25\textwidth,height=0.25\textheight,keepaspectratio]{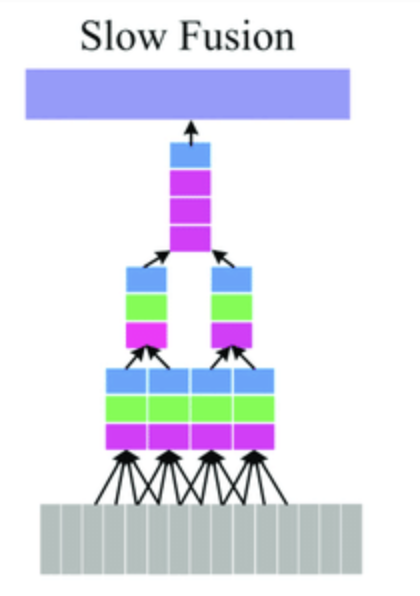}
      \captionof{figure}{\large 3DCNN or Slow Fusion}\label{Fig:sf} 
  \end{figure}

The model receives a four-dimensional tensor of shape H x W x C x T (two spatial dimensions, one channel dimension, and one temporal dimension), enabling it to readily learn all kinds of temporal interactions between adjacent frames\cite{ji20123d,karpathy2014large}.

A disadvantage of this approach is that increasing the input dimensions significantly increases the computational and memory requirements.

 \item {\textbf{Two Stream Models}}
 
A significant problem for many applications, including surveillance, personal assistance, autonomous driving, etc., is understanding and recognising video content. Convolutional neural networks are being used to extract features as part of the current machine learning trend in the field (CNNs). The typical method is to use CNNs on a series of frames, followed by an aggregate over time, or by using CNNs with a spatial-temporal architecture \cite{simonyan2014two}.

 \item{\textbf{Using Optical Flow and CNN’s}}
 
\begin{figure}[h]
    \centering
    \includegraphics[width=1.0\textwidth,height=1.0\textheight,keepaspectratio]{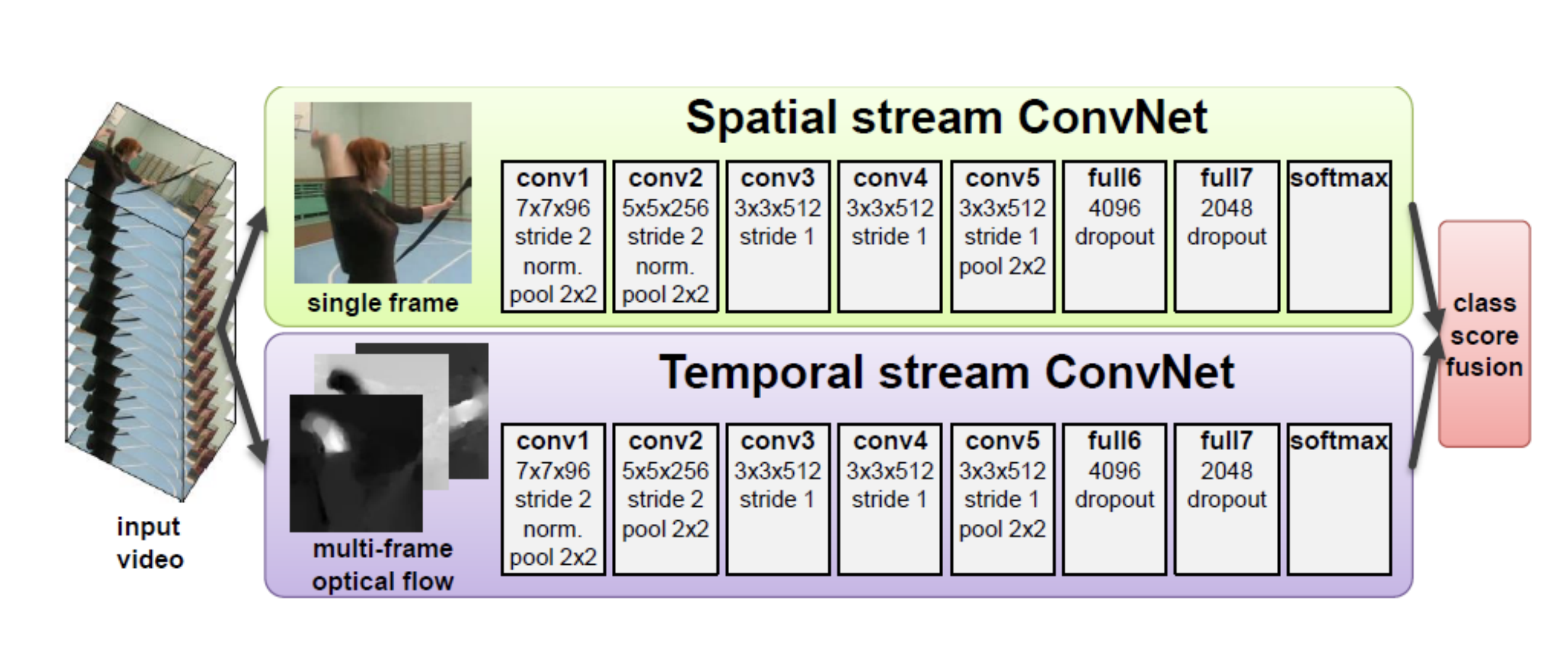}
      \captionof{figure}{\large Two Stream CNN Architecture \cite{sargano2017comprehensive}}\label{Fig:of} 
  \end{figure}
The pattern of apparent motion of objects and edges, known as optical flow, is used to determine the motion vector for each pixel in a video frame \cite{sargano2017comprehensive}.

Two convolutional network streams as shown in Figure \ref{Fig:of} are used in parallel in this method. Spatial Stream is the name of the steam on top. It runs several CNN kernels on a single frame from the video before making a prediction based on the spatial information it contains.

The stream at the bottom, referred known as the Temporal stream, collects the optical flows from every subsequent frame after combining them with the early fusion technique and then uses the motion data to anticipate. Finally, the final probabilities are calculated by averaging the two anticipated probabilities.

This method is flawed because it searches for optical flows for each video using a separate optical flow algorithm from the main network.

 \item{\textbf{SlowFast Networks for Video Recognition}}
 
Using an innovative approach, SlowFast \cite{feichtenhofer2019slowfast} research from Facebook AI Research achieved cutting-edge scores on the prominent video understanding benchmarks Kinetics-400 and AVA. The method's core involves running two concurrent Fast and Slow convolution neural networks (CNNs) on the same video clip. \\ \\
Two independent pathways were already included in earlier methods, such as the Two-Stream method \cite{simonyan2014two}. However, the spatial stream followed one path, whereas the temporal stream followed a different one. As a result, the model had trouble capturing fine motion. \\ \\
This study's essential contribution was to re-conciliate the spatial and temporal streams by supplying raw video to each path at varied temporal rates.
\begin{itemize}
    \item In the SlowPath, a few sparse frames should be able to collect spatial semantic information.
    \item In the FastPath, a high temporal resolution should record Fast and minute motion.
\end{itemize}
Both Pathways execute 3D Convolution operations and employ 3D ResNet. Convolution is performed across an external temporal dimension in addition to channel and spatial dimensions in 3D convolution, an extension of conventional 2D convolution.

\begin{figure}[h]
    \centering
    \includegraphics[width=1.0\textwidth,height=1.0\textheight,keepaspectratio]{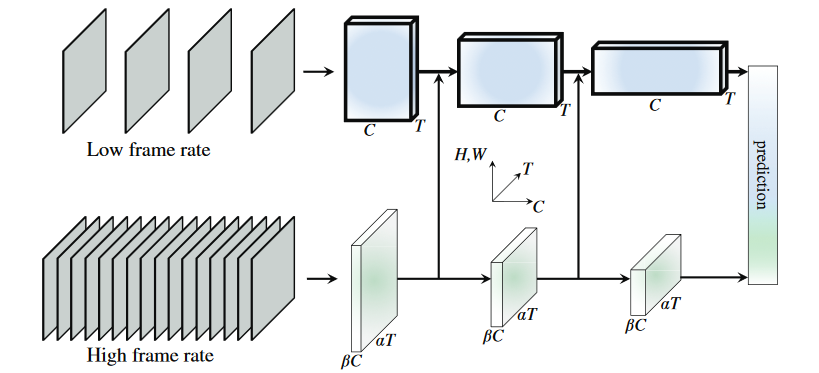}
      \captionof{figure}{\large SlowFast Network \cite{feichtenhofer2019slowfast}}\label{Fig:slow_fast} 
  \end{figure}

The authors' contribution also included introducing three distinct fusing techniques for summarising or concatenating features from the FastPath with those from the SlowPath at various scales.
These are the techniques:
\begin{itemize}
    \item Time-to-channel: All frames should fit inside the channel's dimension.
    \item Time-stride sampling: Choose one frame at a time.
    \item Convolution over time: Execute a 3D convolution with a fixed stride.
\end{itemize}
The fast pathway and the slow pathway are connected laterally.
Thus, combining motion and semantic information can enhance the model's performance. The authors applied a global average pooling layer at the end of each pathway, which concatenates the output from both pathways and decreases dimensionally. Finally, a fully connected layer with the softmax function classifies the action performed in the input clip.

 \item{\textbf{Temporal Shift Module(TSM)}}
\label{sec:temporal_shift_pre}

Conventional 2D CNNs cannot capture temporal correlations when a video is utilised as input. Although 3D CNN-based techniques are frequently employed for video interpretation, they are computationally expensive to implement, especially on embedded devices. While maintaining the complexity of 2D CNN, Temporal Shift Module (TSM) \cite{lin2019tsm} with shift operation can attain 3D CNN's accuracy.

Understanding videos requires the use of temporal modelling. For instance, reversing the video clip will have the opposite results, making it possible to discern between opening and closing the cap of a water bottle.

The entire video is stored in the model, allowing TSM to move previous and subsequent frames with the current frame. However, while running a real-time video, TSM combines the previous and current frames. The illustration can be seen in Figure \ref{Fig:tsm2}. TSM comes in two flavours: residual and in-place.

 \begin{figure}[h]
    \centering
    \includegraphics[width=0.8\textwidth,height=0.8\textheight,keepaspectratio]{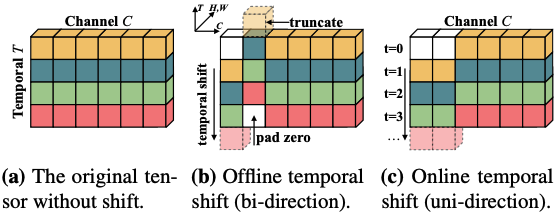}
      \captionof{figure}{\large Temporal Shift Module(TSM) }\label{Fig:tsm1} 
  \end{figure}

When an in-place shift is involved, we include a shift module before each convolutional layer. However, this solution reduces the backbone model's capacity to learn spatial features, particularly when shifting a lot of channels, as the data in the moved channels is lost for the current frame.

\begin{figure}[h]
    \centering
    \includegraphics[width=0.8\textwidth,height=0.8\textheight,keepaspectratio]{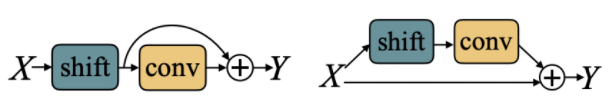}
    \captionof{figure}{\large In-place TSM(left), Residual TSM(Right)}\label{Fig:tsm2} 
  \end{figure}

On the other hand, the problem with the in-place shift solution can be resolved by including a shift module inside the residual branch. Through identity mapping and residual implementation, all the data from the initial activation is still accessible following a temporal shift.

\begin{figure}[h]
    \centering
    \includegraphics[width=1.0\textwidth,height=1.0\textheight,keepaspectratio]{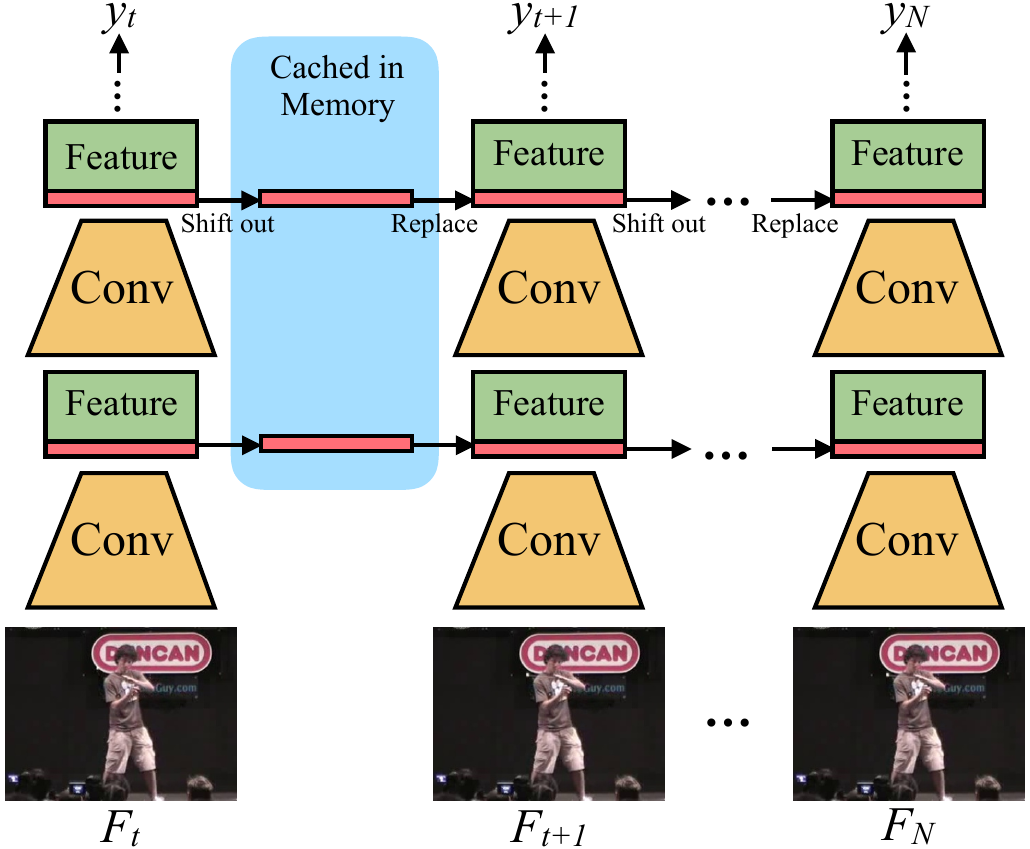}
    \captionof{figure}{\large Uni-directional TSM}\label{Fig:tsm3} 
  \end{figure}
Another conclusion is that the proportion of shifted channels is proportional to performance:
\begin{enumerate}
    \item The ability to handle complex temporal interactions may not be sufficient if the proportion is too small.
    \item The ability to learn spatial features may suffer if the fraction is too high.

\end{enumerate}
Figure \ref{Fig:tsm3} depicts the inference graph of a unidirectional TSM for online video recognition, with a repeat coming after the pipeline. First, each residual block's first 1/8 feature maps are stored and cached in memory for each frame. Second, the cached feature maps are used for the first 1/8 of the current feature maps in the next frame. The final layer produces a 7/8 current and 1/8 historical feature map combination. We can gain several unique benefits by utilising the unidirectional TSM for online video recognition, including:

\begin{enumerate}
    \item Low latency inference: We can replace and cache 1/8 of the features per frame without adding any more computations. As a result, the latency for providing per-frame prediction is nearly identical to the 2D CNN baseline.
    \item Low memory usage: Memory usage is low because just a tiny subset of the features are cached in memory. 
    \item Multiple layers of temporal fusion are possible thanks to TSM.
\end{enumerate}

\end{itemize}

\subsection{Generative Adversarial Networks (GAN)}
\label{sec:gan_lit}

Generative adversarial networks are a type of unsupervised generative model that employs deep learning and generative modelling methods to find and learn patterns in the input data samples. Such a model might be used to generate new, realistic samples that appear to be drawn from the original dataset. 

\subsubsection{Generative Models}
\label{sec:generative_lit}
Statistically, there are two types of models -- The generative type and the discriminative type. The discriminative models largely rely on the conditional probability of an event happening based on a posterior condition. It consists of models such as logistical regression and classifiers. The generative models learn the implicit distribution of the dataset through probability and likelihood estimation. It can generate original synthetic data representations that fit the distribution. Some common generative models are listed below.

\begin{enumerate}
    \item \textbf{Bayesian Network} \\
    It is a graphical generative model based on modelling the probabilistic distribution through a Directed Acyclic Graph (DAG).

    \item \textbf{Generative Adversarial Networks} \\
    These are one of the most famous generative models. They consist of two parts -- the generator and the discriminator, which are trained adversarially to learn the implicit correlations in the dataset. We explain more about these in the upcoming text.

    \item \textbf{Gaussian Mixture Model} \\
    GMM is a probabilistic generative model. It makes an assumption that all data points are formed by a combination of finite gaussian distributions.

    \item \textbf{Hidden Markov Model} \\
    HMM is a statistical model, it is extensively used to find out the correlations between events. It is capable of modelling the evolution of the event.
\end{enumerate}

\begin{figure}[h]
    \centering
    \includegraphics[width=0.8\textwidth,height=0.8\textheight,keepaspectratio]{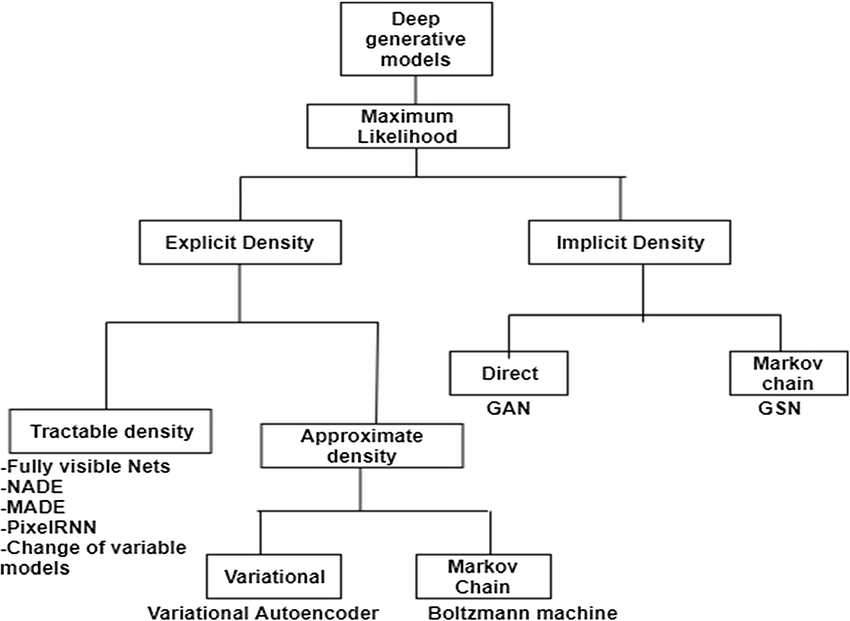}
      \captionof{figure}{\large Taxonomy of Generative Models \cite{yinka2020review}}\label{Fig:taxonomy_gen} 
  \end{figure}

Figure \ref{Fig:taxonomy_gen} shows the taxonomy of the generative models.

\subsubsection{Adversarial Learning}

GANs can be used to estimate the likelihood function through an adversarial process in which two models are trained at the same time: a generative model $\mathcal{G}$ which captures the data distribution, and a discriminative model $\mathcal{D}$ which approximates the likelihood that a sample came from the training data instead of the generative model $\mathcal{G}$. \\
The purpose of $\mathcal{G}$’s training is to make $\mathcal{D}$ more likely to make a mistake. $\mathcal{D}$ is trained in such a way that it should optimize the likelihood of correctly labelling both training and generated samples from $\mathcal{G}$. The Discriminator is seeking to reduce its reward V($\mathcal{D}$, $\mathcal{G}$) in the min-max game that the GANs are designed as, while the Generator is aiming to maximize its loss by minimizing the Discriminator's reward. The following equation mathematically explains it:\\ \\
\begin{equation}
   \min_{G}\max_{D}\mathbb{E}_{x\sim p_{\text{data}}(x)}[\log{D(x)}] +  \mathbb{E}_{z\sim p_{\text{z}}(z)}[1 - \log{D(G(z))}] 
\end{equation}
\\
Where $\mathbb{E}_{x\sim p}(\mathit{f})$ defined the expected value of given function $\mathit{f}$ over the data-points $\mathit{x}$ sampled from the distribution $\mathit{p}$. \\ \\
Let us go through some of the GAN models in the following sub-sections.

\subsubsection{Types of GANs}
This section includes a quick introduction of the most commonly utilised GAN designs, as well as an understanding of their benefits and drawbacks.

\begin{itemize}
\item {\textbf{Vanilla GANs}

The simplest GAN architecture -- The Generator and Discriminator in this scenario are straightforward multi-layer perceptrons. Simple stochastic gradient descent is used in vanilla GAN's approach to optimising the mathematical problem. However, since it uses a feedforward neural network rather than a convolutional neural network to extract features from an image, vanilla GAN designs do not support any true spatial reasoning. Figure \ref{Fig:vanilla_GAN_architecture} shows the basic architecture of vanilla GAN.
\begin{figure}[h]
    \centering
    \includegraphics[width=1.0\textwidth,height=0.4\textheight,keepaspectratio]{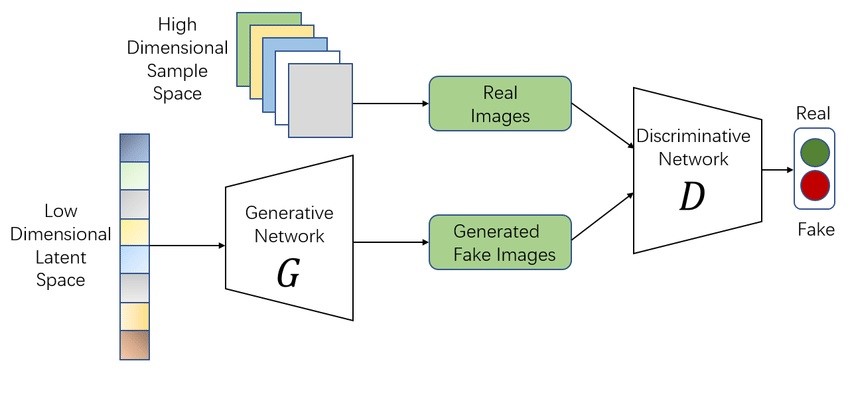}
      \captionof{figure}{\large vanilla GAN architecture}\label{Fig:vanilla_GAN_architecture} 
  \end{figure}
}

\item {\textbf{Deep Convolution GANs (DCGANs)}

In their study, Radford et al. \cite{radford2015unsupervised}, shows that Unsupervised Representation Learning With Deep Convolutional Generative Adversarial Networks opens up whole new frontiers for generative models.
The generator and discriminator use the convolution layers along with the batch normalisation, and the LeakyRelu layer. The generator is able to learn the implicit distribution for computer vision datasets, which was unachievable using only the Dense layers.
\begin{figure}[h]
    \centering
    \includegraphics[width=1.0\textwidth,height=0.4\textheight,keepaspectratio]{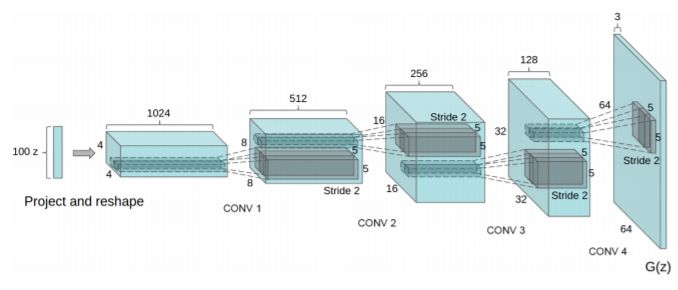}
      \captionof{figure}{\large DCGAN \cite{radford2015unsupervised}}\label{Fig: DCGAN} 
  \end{figure}
}

\item {\textbf{Conditional GANs (CGANs)}

The model takes a long time to converge and, in many instances, keeps oscillating and never converging, which is one of the most frequent issues with classic GANs. Furthermore, the nature of the created data is uncontrollable. The GAN architecture has seen a number of revisions since 2014. CGANs are one such enhancement. We may overcome the aforementioned issues by using conditional GANs, which are an extension of standard GANs.\\ \\
They are generative adversarial networks that were first introduced in 2014 by Mirza et al. \cite{mirza2014conditional}, their Generator and Discriminator is trained to utilise extra data. This might be the image's class or a list of specific attributes we want the output to have. Including the property information modifies the output and tells the generator to create the desired result. The discriminator's role in Conditional GAN involves not just telling real data from predicted data but also determining if predicted data agrees with the supplied information. The advantage of utilising Conditional GAN is that convergence will happen more quickly, the output generated will be more relevant to the input than being fully random, and the discriminator will be better able to tell the difference between actual and created data. \\ \\
Both CGAN and ordinary GAN have the same loss functions. When the discriminator's duty is relatively simple in the early stages of GAN training, the min-max loss function might lead the GAN to become stuck. Additionally, there is a significant problem with vanishing gradients when the discriminator is perfect, causing the loss to be zero, and there is no gradient to update the weights during model learning. The fact that CGANs are not entirely unsupervised and require labels to operate is another drawback.}

\item {\textbf{Wasserstein GANs (WGANs)}

\textbf{Problem of Mode Collapse: }A skilled generator should be able to provide a wide range of outputs. When the Generator is in a mode collapse state, it can only create a single output or a limited number of outputs, regardless of the input. This could occur as a result of different training problems, such as the generator discovering data that can trick the discriminator and continuing to produce such data. \\ \\
\textbf{Problem of Vanishing Gradient: }When a deep multi-layer feed-forward network is unable to transmit relevant gradient data from the model's output end back to its input end, the gradient is said to be "vanishing." The model might not be adequately trained as a result, and it might converge too quickly to a subpar answer. This issue arises because the gradient keeps growing less and smaller as it flows backwards. It can grow so tiny that the first few layers (at the input end) either learn extremely slowly or not at all. As a result, weights won't be updated, and the model's overall training will come to an end.\\ \\
The aforementioned issues are present in both conventional GAN and CGAN. Wasserstein GAN was created in 2017 in \cite{arjovsky2017wasserstein} to address these problems. The model may be trained more steadily with WGAN. Additionally, it offers a more accurate representation of the data distribution seen in a particular training dataset. WGAN employs the critic that measures the realness or fakeness of the picture, in contrast to the conventional GAN, where we use a discriminator to forecast if an image is genuine or fake. This modification is based on the notion that reducing the disparity between the data distributions of training data and produced data should be the primary goal of training the generator. \\ \\
Instead of employing Binary Cross-Entropy(BCE) loss in Wasserstein GAN, we utilise Wasserstein loss based on Earth-distance. The Earth-Mover’s distance is a measure of distance between two probability distributions over some region D. The EM(Earth-Mover’s) distance is continuous and differentiable, which means that the critic can be trained to optimality. The generator would have to attempt something new if the discriminator did not become stuck on local minima and reject the output that it stabilises on. Therefore, Wasserstein GAN is used to address the mode collapse problem. The issue of vanishing gradients is also overcome since the EM distance is differentiable. We utilised weight clipping since the critic must meet the 1-Lipschitz restriction. Following is the algorithm \cite{arjovsky2017wasserstein}.\\
\begin{figure}[h]
    \centering
    \includegraphics[width=0.9\textwidth,height=0.9\textheight,keepaspectratio]{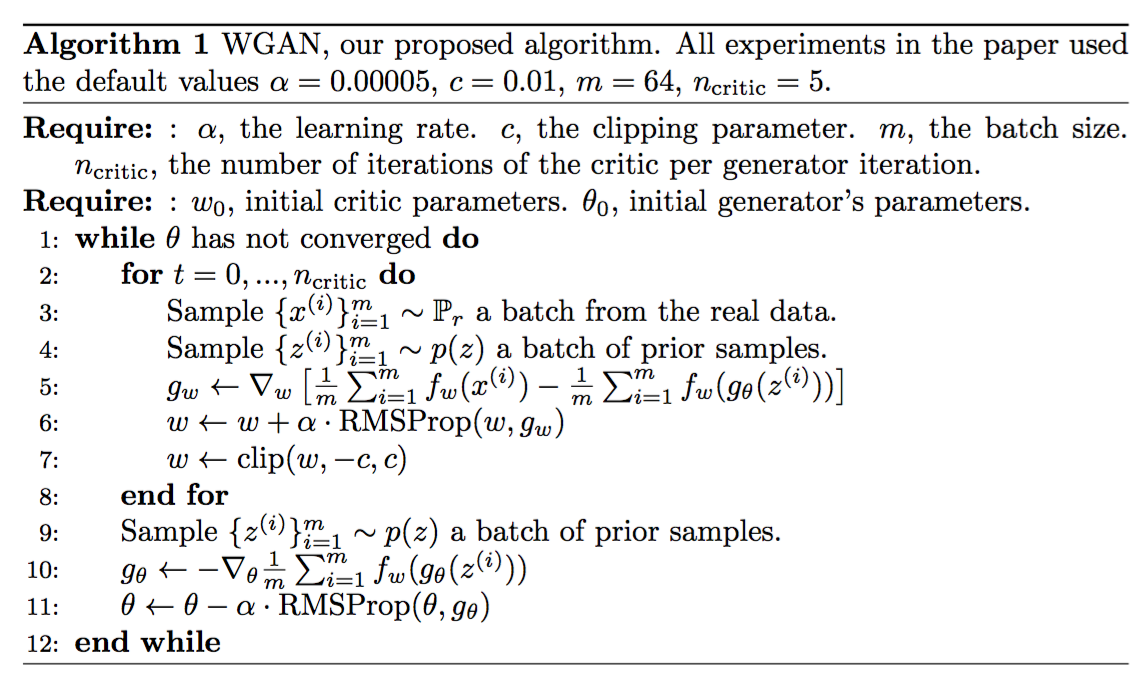}
      \label{Fig: WGAN} 
  \end{figure} \\
Although it has been shown that training in WGAN is slower than in standard GAN, the former is still preferable because of its many benefits, including increased stability and the elimination of issues like mode collapse and vanishing gradient.}

\end{itemize}

\subsubsection{Image-to-Image Translation}
\textit{Image-to-Picture Translation} \cite{dong2017unsupervised} refers to the automatic transformation of an image's original form into various synthetic forms (style, partial contents, etc.) while preserving the semantics or structural integrity of the original image. \\
In this study, we concentrate on converting photographs from one domain to another, such as changing the gender or the faces. A general approach to achieve “Image-to-Image Translation” by using DCNN and cGAN. The authors create a two-step unsupervised learning technique in this study to interpret images without defining a relationship between them.

The entire network architecture is shown in Figure \ref{Fig:i2i1}.

\begin{figure}[h]
    \centering
    \includegraphics[width=1.0\textwidth,height=0.4\textheight,keepaspectratio]{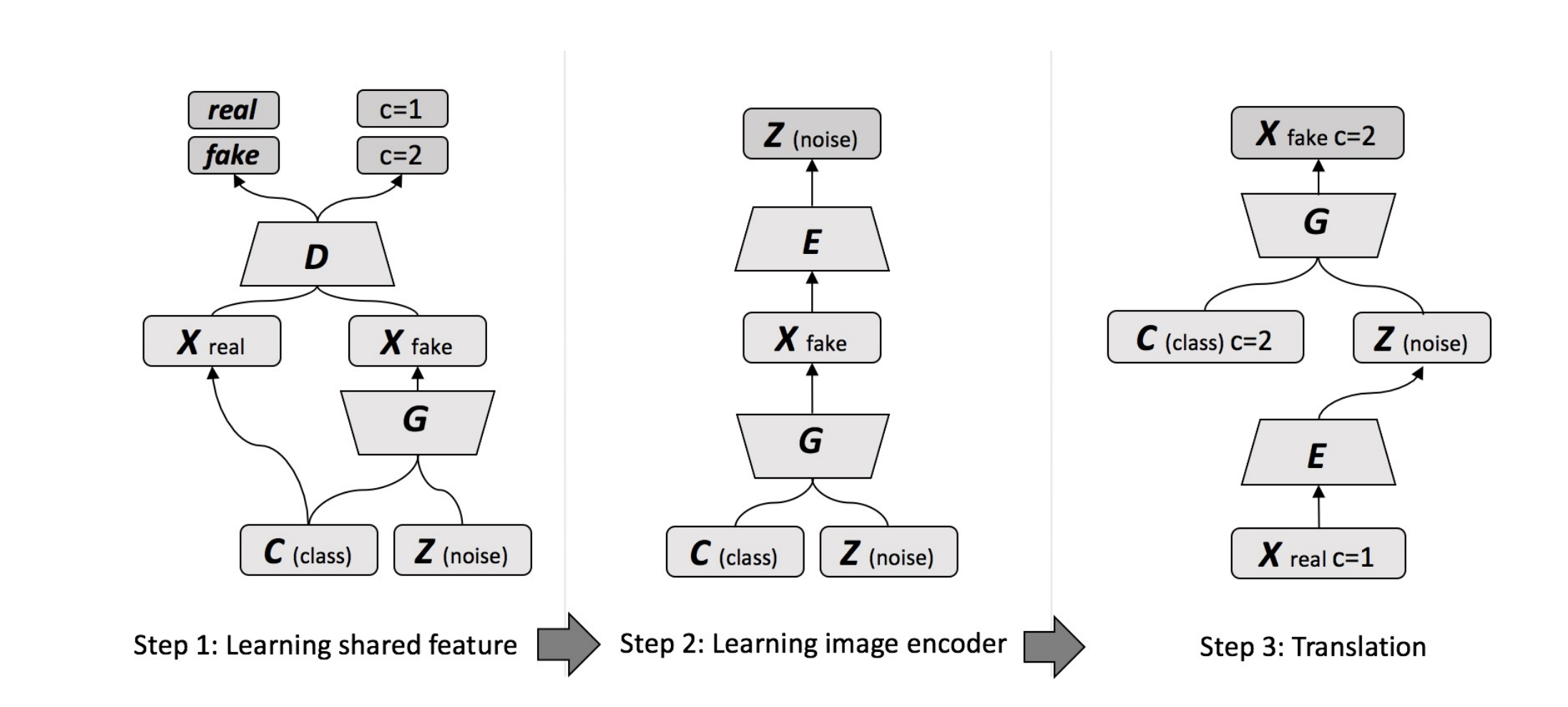}
      \captionof{figure}{\large Image-To-Image Translation \cite{dong2017unsupervised}}\label{Fig:i2i1} 
  \end{figure}

It's a two-step learning process: learning shared features and learning image encoder.

\begin{enumerate}
    \item \textbf{Learning Shared Feature: }According to the left side of Figure \ref{Fig:i2i1}, To discover the general shared qualities of samples of images drawn from various domains, it makes use of an auxiliary classifier. A latent vector z is used to represent these shared characteristics. By keeping the latent vector unchanged and altering the class label after this step, generator G can produce corresponding images for other domains.
    \item \textbf{Learning Image Encoder: }To embed images into latent vector. After generating generator G in the first phase, they use image encoder E and train it by minimizing the MSE between the input latent vector and output latent vector, as indicated in the center of Figure \ref{Fig:i2i1}.
    \item \textbf{Translation: }Following the aforementioned two phases, as illustrated in the right column of Figure \ref{Fig:i2i1}, images can be translated using trained E and trained G. They embed X real with domain/class label c=1 into latent vector Z using the learned image encoder E given an input image X that needs to be translated. The trained generator G will then output the false image X fake with the input Z and another domain/class label c=2.
\end{enumerate}

Recently, certain GAN models have been introduced to accomplish the above task. These models \cite{isola2017image} have shown promising results and are being intensively researched today. Here is an overview of the recent breakthroughs in image-to-image transformation:
\begin{itemize}

\begin{figure}[h]
    \centering
    \includegraphics[width=1.0\textwidth,height=0.3\textheight,keepaspectratio]{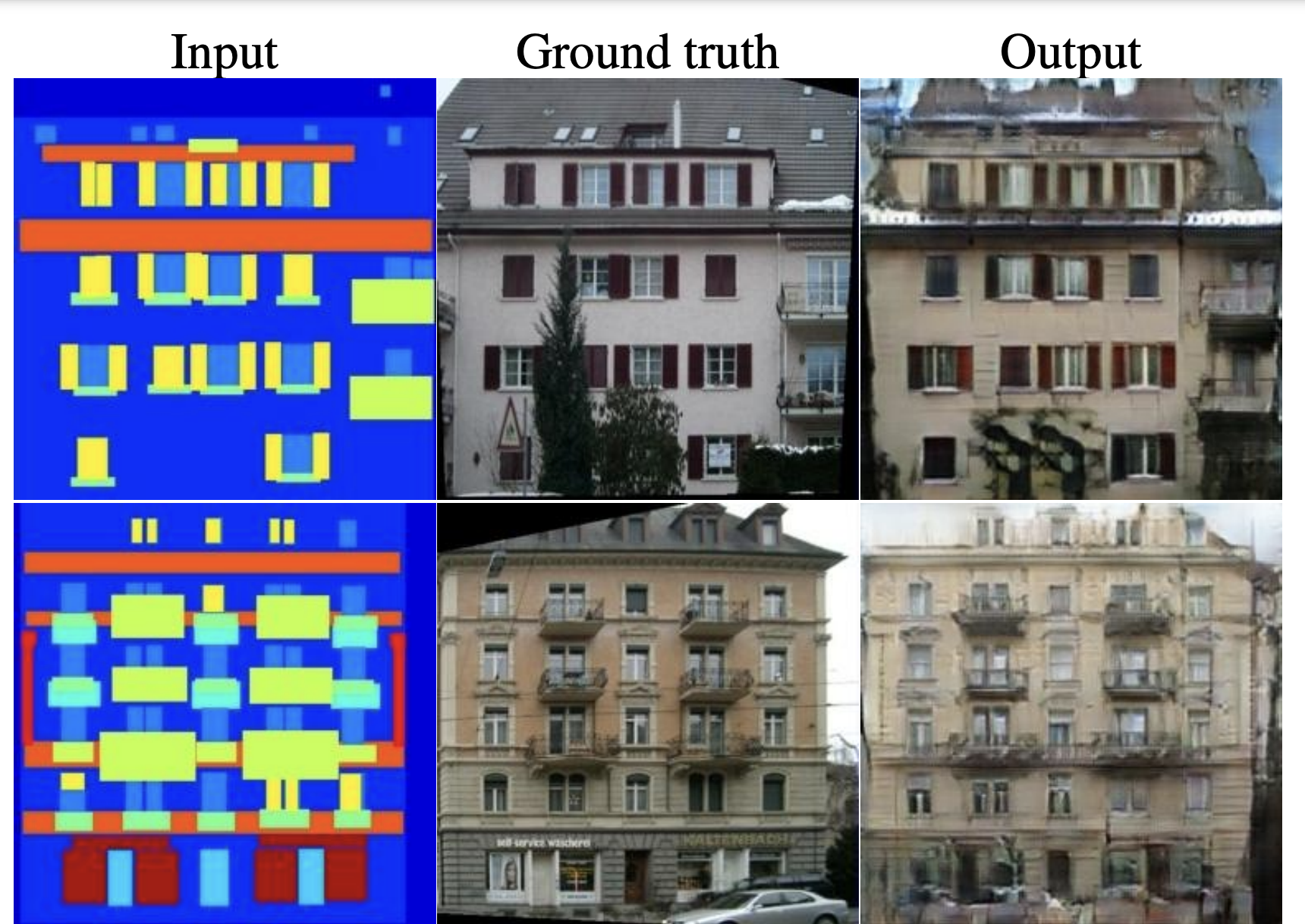}
      \captionof{figure}{\large Pix2Pix is a conditional GAN \cite{isola2017image}}\label{Fig:p2p1} 
  \end{figure}
  
\item\textbf{{Pix2Pix}}

Phillip Isola et al. \cite{isola2017image}. created the conditional GAN (cGAN) known as Pix2Pix. Contrary to vanilla GAN, which uses real data and noise to learn and produce images, cGAN produces images using real data, noise, and labels.

Essentially, the generator picks up the mapping from the noise and the true data.
$$
G:\{x, z\} \rightarrow y
$$

The discriminator similarly gains representational knowledge from labels and actual data.

$$
D(x, y)
$$

With this configuration, cGAN can be used for image-to-image translation tasks where the generator needs an input image to produce an output image that matches. In other words, the generator generates a target image by utilising a condition distribution (or data) such as instruction or a blueprint (see the Figure \ref{Fig:p2p1}).

  \begin{figure}[h]
    \centering
    \includegraphics[width=1.0\textwidth,height=1.0\textheight,keepaspectratio]{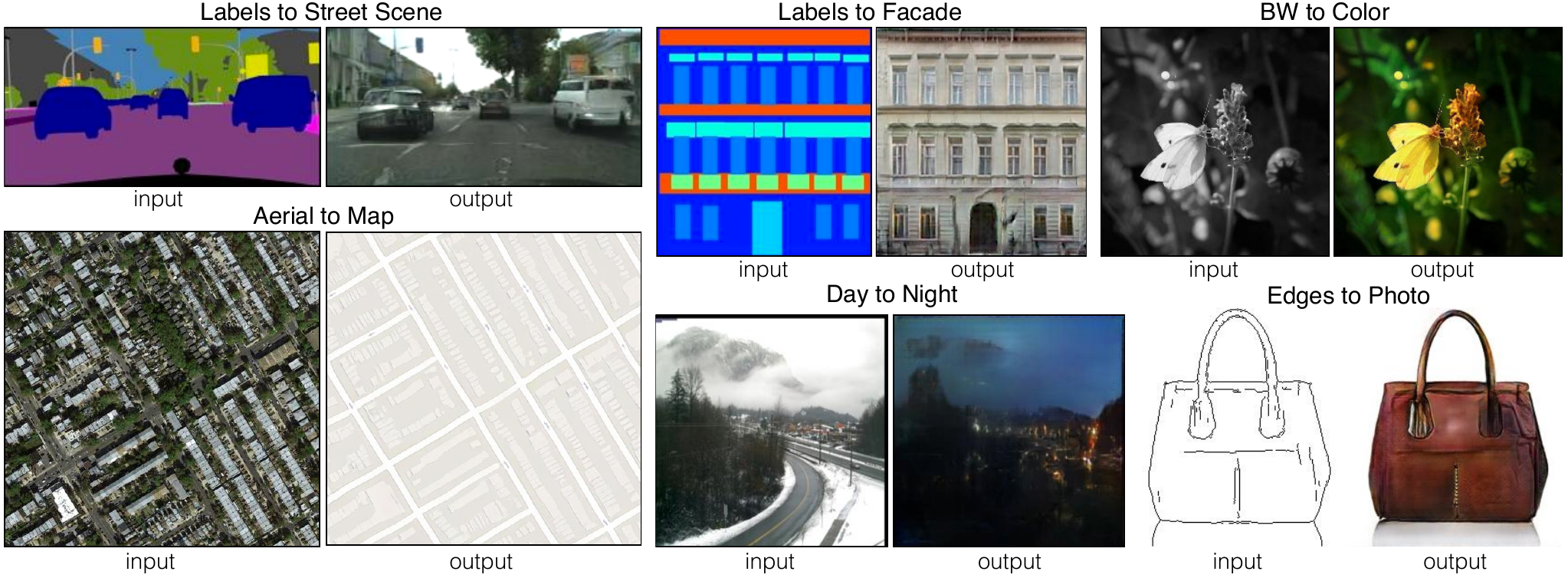}
      \captionof{figure}{\large Application of Pix2Pix \cite{isola2017image}}\label{Fig:p2p2} 
  \end{figure}

Pix2Pix, which stands for \textit{pixel-to-pixel}, is a platform with several different techniques for analysing and interpreting original content. Pix2Pix can turn sketches and illustrations into paintings using artificial intelligence, machine learning, and Conditional Adversarial Networks (CAN). The authors provide an online interface for the pretrained pix2pix model. You can enter a picture or do a quick drawing in the input box and click convert to let Pix2Pix turn it into a painting.

\item\textbf{{CycleGAN}}

The cyclic structure created between these several generators and discriminators gives CycleGAN its name.Pix2Pix is a fantastic model, but it can't be used without an image-to-image or one-to-one dataset. However, a model called CycleGAN can address this issue. 

\begin{figure}[h]
    \centering
    \includegraphics[width=1.0\textwidth,height=1.0\textheight,keepaspectratio]{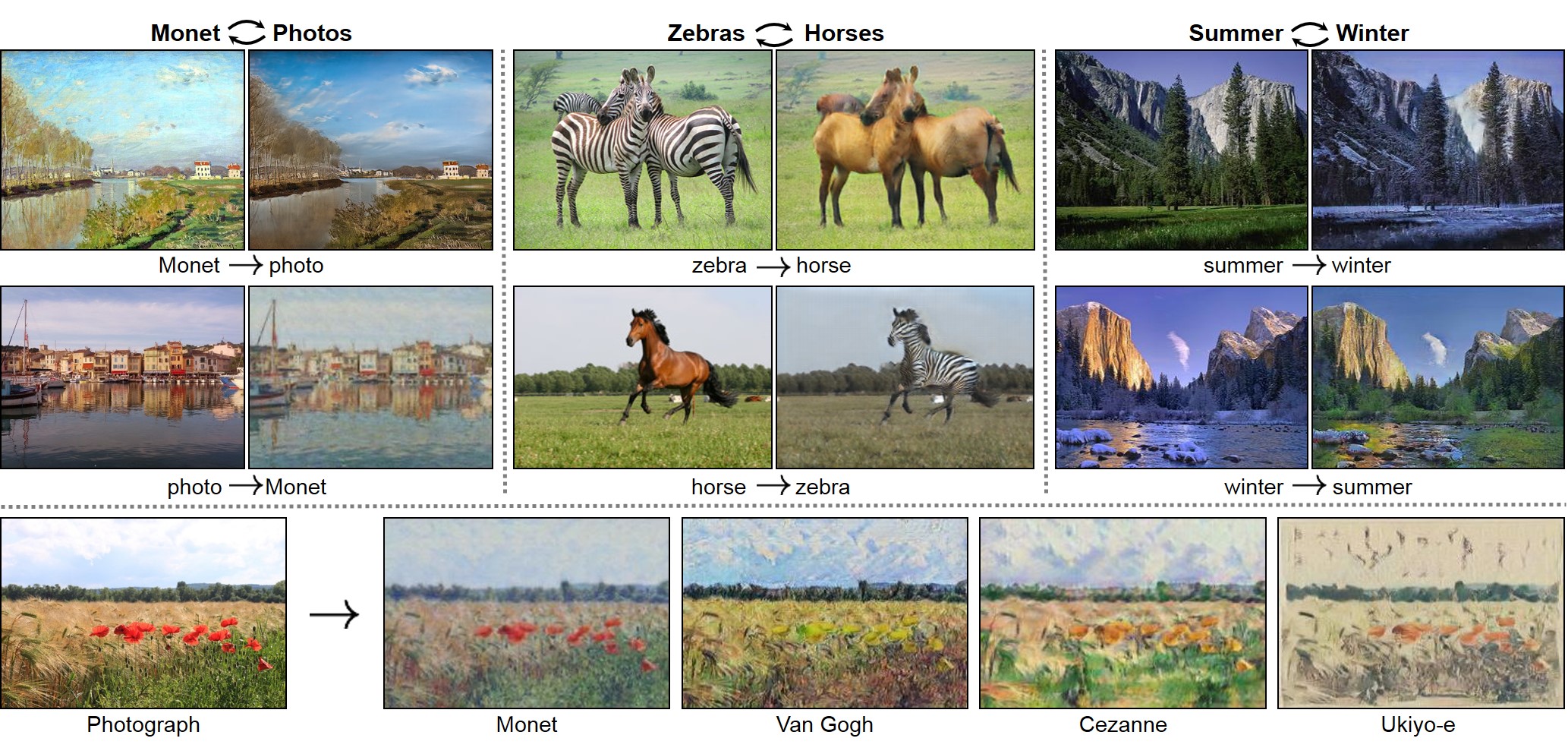}
      \captionof{figure}{\large CycleGAN \cite{isola2017image}}\label{Fig:cg} 
  \end{figure}

CycleGAN has the ability to transfer styles to photos without requiring a one-to-one mapping between the images, which is a clear advantage. Image translation can be integrated with a variety of ways using CycleGAN. Zebras can be transformed into horses, or you can alter summer images into winter ones in Monet's paintings and photography. It also changes both ways.
Refer to Figure. \ref{Fig:cg}

Assume that A is summer and B is winter. Now, the cyclic flow appears to be as follows:

\begin{itemize}
    \item Generator A-to-B (Summer\( \longrightarrow \)Winter) is fed random samples from domain A (Summer). This generator takes photos from Summer and converts them to Winter. So, instead of using random noise (like in normal GAN), we train the generator to translate images from one domain to another.
    \item The second generator, generator B-to-A (Winter\( \longrightarrow \)Summer), receives this created picture for Domain B (Winter) from generator A-to-B, repeating the input image from Domain A.
\end{itemize}

The same flow then goes vice-versa for Winter →Summer.

\vspace{0.5cm}
\item\textbf{{StarGAN}}

While CycleGAN and Pix2Pix have successfully translated images, they hit a wall when dealing with more than two domains.
A generative adversarial network called StarGAN can learn mappings between several domains. 

\begin{figure}[h]
    \centering
    \includegraphics[width=0.7\textwidth,height=0.7\textheight,keepaspectratio]{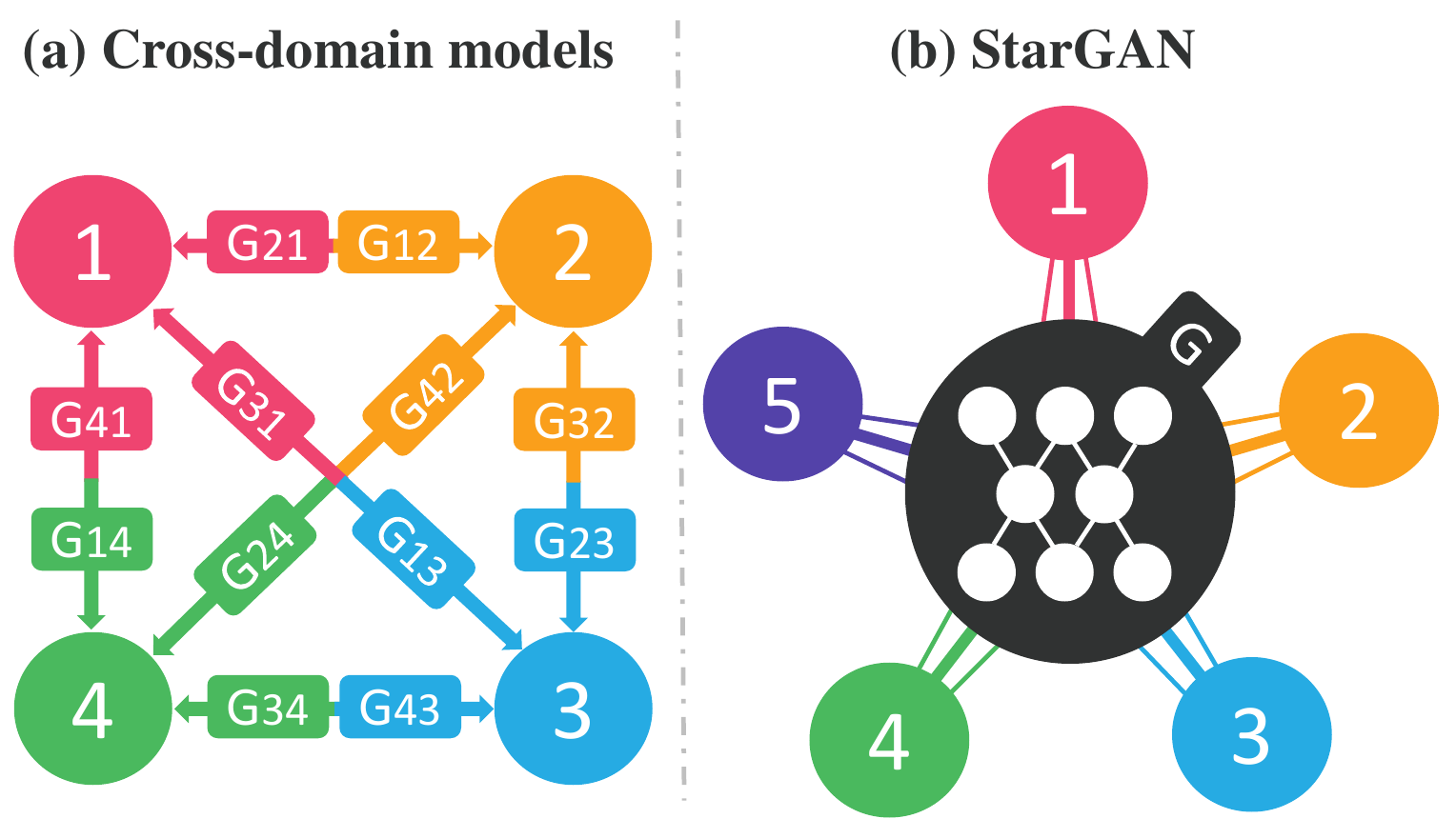}
      \captionof{figure}{\large Cross Domain v/s StarGAN \cite{choi2018stargan}}\label{Fig:c/s} 
  \end{figure}
  
The paper that introduced StarGAN was titled \textbf{Unified Generative Adversarial Networks for Multi-Domain Image-to-Image Translation \cite{choi2018stargan}}. Let's break that title down — the keywords being \textit{unified} and \textit{multi-domain}. Traditionally, we would train a model and create new images using the labels from the provided image distribution. The CELEBA dataset, for instance, contains 40 labels, some of which are Pointy Nose, Wavy Hair, Eye Glasses, Bangs, Smiling, Oval Face etc.

If in case we need to add a new label to our dataset, we need to transfer the latent space from a different data set containing the relevant attribute to the generator tuned to CelebA. Before StarGAN was suggested, two separate GANs were trained to accomplish this. Figure \ref{Fig:c/s} shows how the models' dependence on one another would grow as the reliance on various datasets increased. If there are N different models, It will require $\frac{N \times (N-1)}{2}$ combinations.
\begin{figure}[h]
    \centering
    \includegraphics[width=0.8\textwidth,height=0.8\textheight,keepaspectratio]{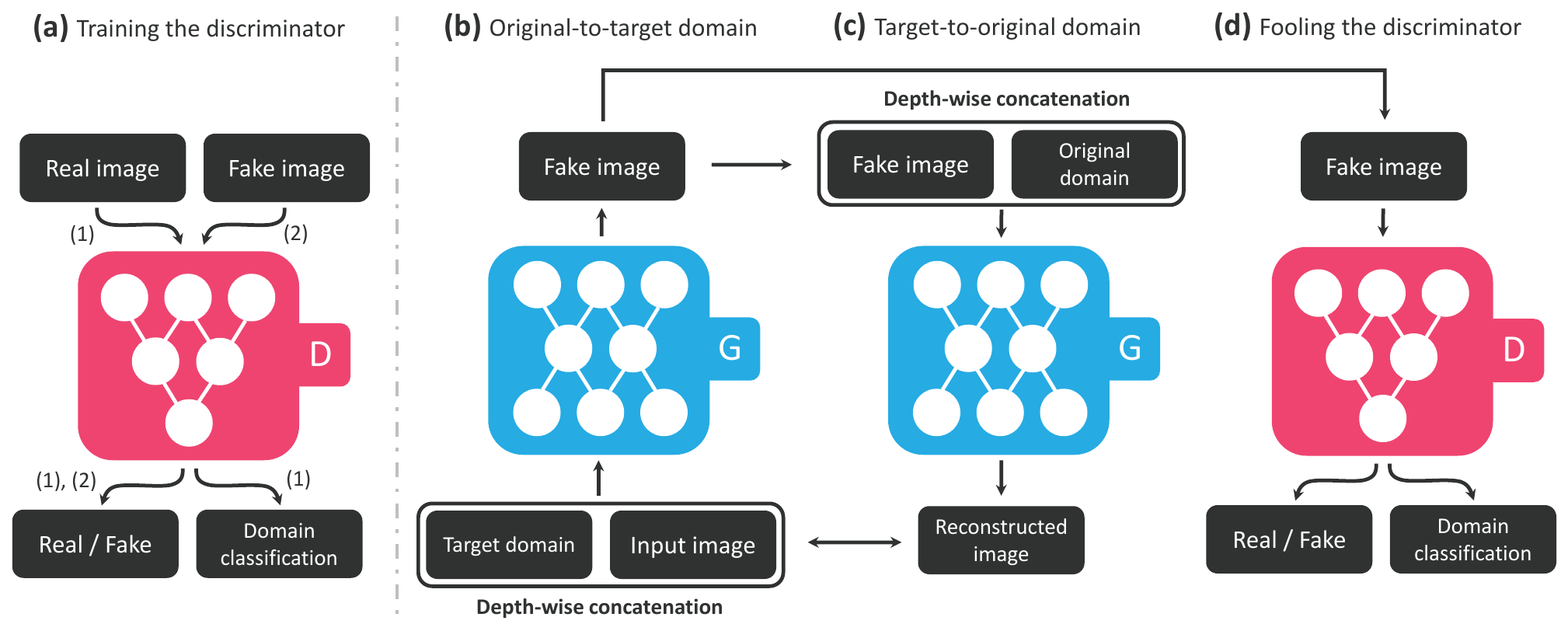}
      \captionof{figure}{\large  Overview of StarGAN \cite{choi2018stargan}}\label{Fig:staro} 
  \end{figure}

Its unified modelling architecture allows it to train numerous datasets and domains simultaneously in a single network—a new, scalable method for performing image-to-image translation among several domains. Any target domain can be translated for the input images. Compared to current techniques, it can produce photos with superior visual quality. It can transfer face characteristics and synthesise facial expressions.

\begin{figure}[h]
    \centering
    \includegraphics[width=1.0\textwidth,height=1.0\textheight,keepaspectratio]{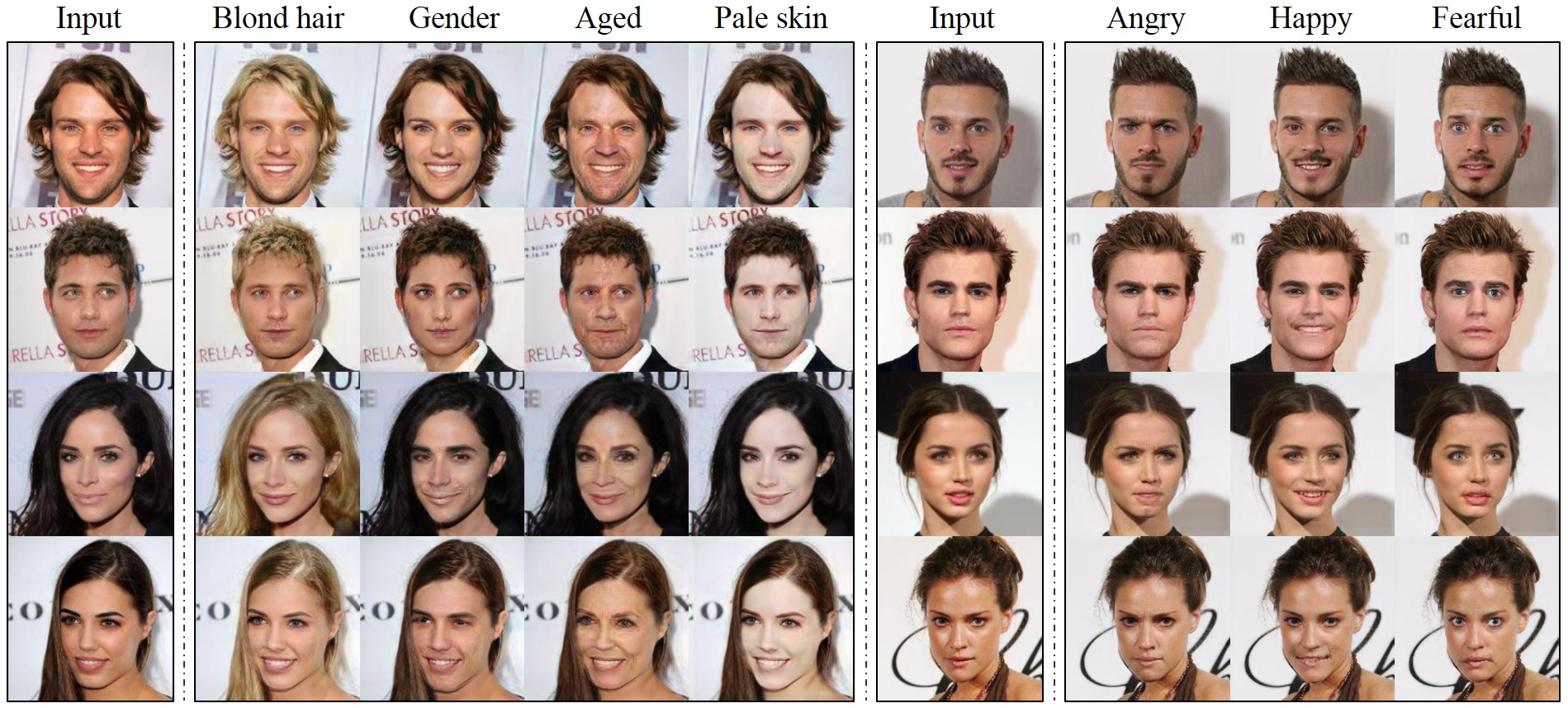}
      \captionof{figure}{\large Multi-attribute transfer results \cite{choi2018stargan}}\label{Fig:matr} 
  \end{figure}

The terms \textit{unified} and \textit{multi-domain} in StarGAN describe the model's main contribution, where Multiple picture datasets can train a single (unified) model, and their feature properties (domain) can be exchanged across them. How does StarGAN accomplish this? It employs a concept known as a mask vector comparable to subnet masks in network theory. To depict the dataset being fed, an additional vector is added to the inputs, and all values except one are turned off. \\ \\

  
Figure \ref{Fig:matr} demonstrates how StarGAN can transform multiple domains such as hair colour, sex, age, complexion, and facial expressions using a single person's shot. All of this is completed simultaneously.
\end{itemize}

\subsection{Anomaly Detection}
\label{sec:anomaly_detection_lit}
\textit{Anomalies are outliers or rare events that are not expected to occur in the scene. A regular event can also be classified as an anomaly if done in an unusual scene.} \\ \\
This is a version of the commonly adopted definition of anomalies. The inherent ambiguity in this formulation comes from the complex relationship between the scene and its constituent objects. A relaxed approach is taken in most studies by taking a suitable definition of outliers.

\subsubsection{Types of Video Anomalies}
Ramachandra et al.\cite{Ramachandra2022ASO} specifies the different types of anomalies in benchmark datasets and practical scenarios.

\begin{enumerate}
    \item {\textbf{Appearance-based Abnormality}} \\
    As the name suggests, the anomalies are based on visual properties like the colour and shape of the object. 

    \item {\textbf{Motion-based Abnormality}} \\
    The anomaly is based on the motion characteristics of the objects. Running and jaywalking are examples of human-centric anomalies. These can be short-term events or long-term abnormal trajectories.

    \item {\textbf{Group Anomalies}} \\
    As the name suggests, a normal event becomes an anomaly if performed together by a lot of constituents. For example, a marching band or a flash mob.  
\end{enumerate}

\subsubsection{Formulations of Anomaly Detection}
There are many papers on Anomaly detection that have given a different formulation of the problem. As per Ramachandra et al. \cite{Ramachandra2022ASO}, there are two major fields of outlier analysis.

\begin{enumerate}
    \item \textbf{The Single-Scene Scenario:} 
    The anomalies are based on the appearance and motion of the foreground objects, as well as the spatial locality of the event. For example, A pedestrian walking on a highway is considered anomalous, whereas walking on a footpath is not. The background scene remains constant throughout the video clip.
    \item \textbf{The Multi-Scene Scenario:} 
    In such scenes, only the appearance and motion of the foreground are taken into account. This complexity arises due to the changing scenes, for example, the background building or the time of the day. The problem has been formulated in \cite{liu2018future, morais2019learning, sultani2018real}. 
\end{enumerate}
In another class of formulations, a training-free method is proposed \cite{del2016discriminative, tudor2017unmasking, zhao2011online}. Such methods predict the deviations in the testing dataset as an alias for anomalies. These are not trained on normal images and function similarly to anomaly predictions in stocks. \\ \\
The most widely accepted formulation of Anomaly detection as presented in \cite{keogh2005hot, liu2018classifier, pang2020self}, involves training the model on a training dataset to learn the normal sample distribution and use it to detect outliers. 

   \begin{figure}[h]
    \centering
    \includegraphics[width=1.0\textwidth,height=0.6\textheight, keepaspectratio]{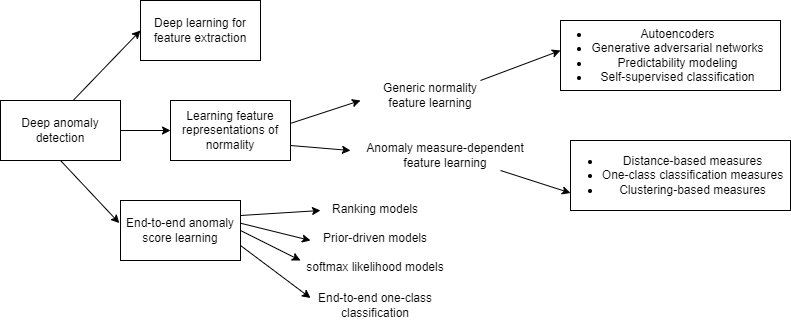}
       \captionof{figure}{\large Anomaly Detection Models }\label{Fig:ad_models} 
   \end{figure}

\subsubsection{Traditional Techniques}
\label{sec:handcrafted_lit}
  
Research in anomaly detection started as early as 1980s. The main goal of these techniques is to describe a video event using manually created features, such as trajectory features \cite{quispe2021trajectory}, low-level features taken from local 3D gradients or dynamic textures, or histograms of oriented gradients (HOG) \cite{hu2018abnormal}, and optical flow (HOF)  \cite{colque2016histograms}. Then, using hand-crafted characteristics linked to typical occurrences, an outlier detection model was built. These approaches are shown in \ref{Fig:trads}. Although these methods worked well for simpler datasets, they could not be adapted to complex ones \cite{adam2008robust}. Further, feature extraction was a time-consuming and labour-intensive procedure, and the hand-crafted features required prior understanding of data; hence this class of algorithms is not appropriate for deciphering complicated video surveillance scenarios.

\begin{figure}[h]
    \centering
    \includegraphics[width=0.8\textwidth,height=0.4\textheight, keepaspectratio]{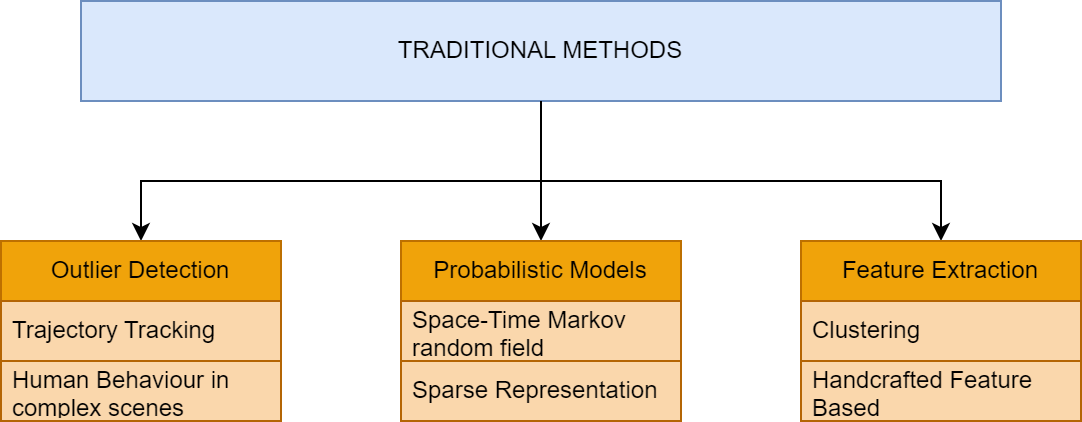}
      \captionof{figure}{\large Traditional Anomaly Detection Methods }\label{Fig:trads} 
  \end{figure}

\vspace{2cm}
Recently the introduction of Deep Learning(DL) has opened new domains for anomaly detection, and Autoencoder(AE) is the DL-based neural network which is the basis for anomaly detection.

\begin{itemize}
\label{sec:autoencoder_lit}

    \item \textbf{Autoencoder (AE)} \\
    Autoencoders(AE) are an unsupervised learning method that uses neural networks to learn representations. In particular, we'll create a neural network architecture that induces a compressed knowledge representation of the original input by imposing a bottleneck on the network. This compression and subsequent reconstruction would be challenging if the input features were not reliant on one another. But let's say the data is structured in some way (i.e. correlations between input features). If so, it is possible to learn this structure and use it to push input through the network's bottleneck.
    
    An AE is used for a specific network architecture to learn effective embeddings of unlabeled input. The AE comprises two parts: an encoder and a decoder. The decoder accomplishes the inverse, converting the latent space back to higher-dimensional space. In contrast, the encoder compresses the data from a higher-dimensional space to a lower-dimensional space (also known as the latent space). By making the decoder output the data fed to it as input, the decoder ensures that latent space can collect most of the information from the dataset space. For the block diagram see the Figure \ref{Fig:ae}
    
    \begin{figure}[h]
        \centering
        \includegraphics[width=1.0\textwidth,height=1.0\textheight,keepaspectratio]{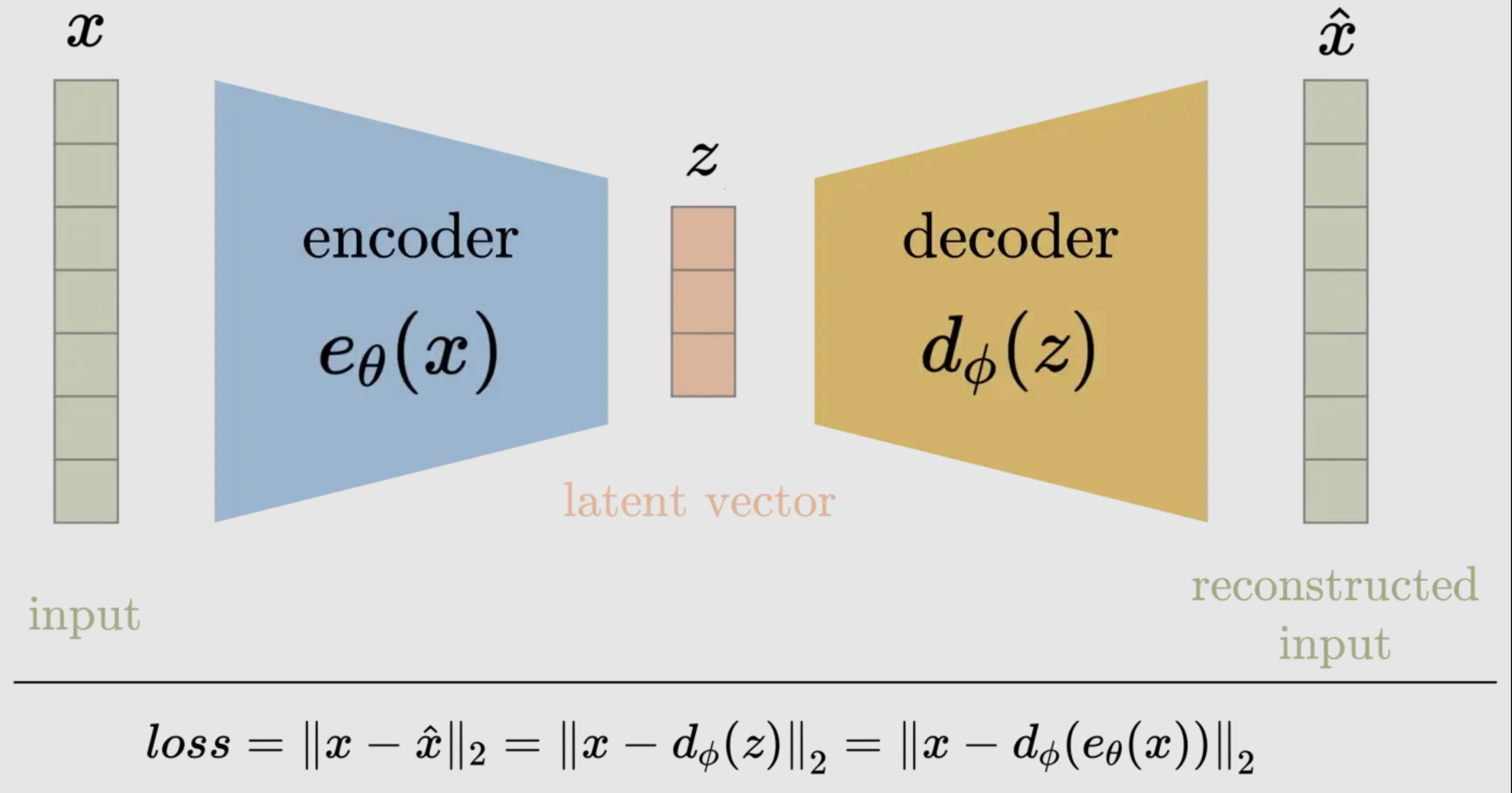}
          \captionof{figure}{\large Autoencoder }\label{Fig:ae} 
      \end{figure}
    
    During training, the encoder function \( e_{\theta}(x) \) is fed the input data x. The input is passed through several layers to create a compressed latent vector z. User-controlled parameters include the number of layers, the kind and size of the layers, and the dimension of the latent space. Compression is achieved if the latent space's dimension is smaller than the input space's, eliminating redundant properties.
    
    The decoder \( d_{\phi}(z) \) typically (but not always) consists of layers nearly complementary to the layers used in the encoder but arranged in the opposite direction. The actions of the original layer, such as transposed Conv layer to Conv layer, pooling to unpooling, etc., can be partially reversed by using a near-complement layer of the layer.
    \begin{itemize}
    
        \item \textbf{Autoencoder for compression}
    A linear AE's latent space closely matches the eigenspace produced by the principal component analysis of the data. The first m eigenvectors of PCA will span the same vector space as a linear AE with input space dimensions of n and latent space dimensions set to mn. \textit{Why use AE if PCA is similar?}
    The ability of AE to train rather potent representations of the input data in lower dimensions with significantly less information loss increases with the addition of nonlinearity, such as nonlinear activation functions, and additional hidden layers.
    
        \item\textbf{Variational Autoencoder}
    The problem of non-regularized latent space in AE is addressed by the Variational AE, which gives the entire space generative capabilities. Latent vectors are output by the encoder in the AE. The VAE encoder produces parameters of a pre-defined distribution in the latent space for each input rather than vectors in the latent space. The latent distribution is then forced to conform to a normal distribution by a constraint placed on it by the VAE. The latent space is ensured to be regularised by this constraint.
    
    \end{itemize}
    AD methods can be divided into two broad approaches. i) The reconstruction based and ii) the prediction based.
\end{itemize}

\subsubsection{Reconstruction Method}
\label{sec:reconstruction_pre}

This method uses an autoencoder architecture to recreate the input sample. The autoencoder consists of an encoder and decoder network. The encoder network down-samples the image into a lower dimensional representation of the image. This latent space consists of a set of compressed features. The decoder learns to reconstruct original-looking images from low-dimensional representations. This is a form of unsupervised approach where the target is the same as the input image. Ref. \cite{hasan2016learning} proposed the use of a 3D CNN-based autoencoder. The autoencoder is trained on the train set consisting of only the normal frames. As a result, the model is able to correctly and accurately reconstruct the output frames $\hat{I}_t$ from the ground truth images $I_t$, whereas it is unable to reconstruct the abnormal frames and the reconstruction error comes out higher. This becomes the basis of anomaly detection using reconstruction.

\subsubsection{Prediction Method}
\label{sec:prediction_pre}

The method has recently been a vivid topic of research for anomaly detection tasks. It makes use of a typical GAN-based architecture comprising a generator and discriminator. For the prediction of the future frame, the model needs to learn the temporal motion information along with the spatial information to give accurate results. Such models typically use some additional network like optical flow, 3D CNN and LSTM models. The predicted images $\hat{I}_{t+1}$ are compared the the ground truth $I_{t+1}$. This approach gives better results as abnormal events usually do not conform to the expected norms.     

\subsubsection{Anomaly Score}
 A model trained on only regular images will be able to reproduce high-quality normal frames but will not be able to reconstruct anomalous images properly. In order to evaluate this visual distinction between reconstructions of real \& fake classes, an anomaly score is introduced.

\begin{figure}[h]
    \centering
    \includegraphics[width=0.60\textwidth,height=0.60\textheight,keepaspectratio]{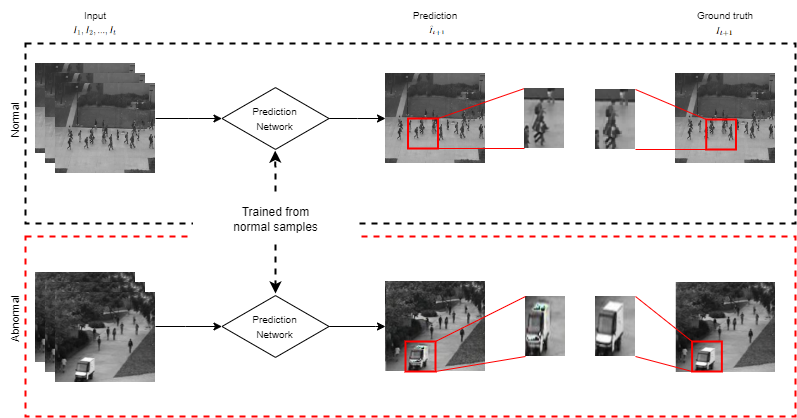}
      \captionof{figure}{\large Anomaly Detection Module }
      \label{Fig:anomaly_module} 
  \end{figure}

Different types of Anomaly Scores have been introduced in prior research work, For example, Mean Squared Error and the Peak Signal Noise ratio, refer to section \ref{methodology}. Further, we employ the thresholding process, as discussed below. 

\begin{itemize}
    \item \textbf{Thresholding} \\
    It is a painstaking task to segregate the abnormal frames from the video. The anomaly score of each frame defines a mathematical standard for measurement of the extent of the abnormality. By observing the distribution of this score, we should decide on a threshold for the anomaly. All the frames above that threshold are labelled as anomalies, and those below are labelled as normal. Using this methodology, we become capable of evaluating other useful metrics. The algorithm \ref{alg:anom_score} is formulated below.
    \begin{algorithm}[H]
\Large
\caption{ \Large \ \ Thresholding Process}\label{alg:anom_score}
\begin{algorithmic}[1]
\Require Ground Truth Frames: $\{ I_1, I_2, ..., I_N \} $, Predicted Frames: $\{ \hat{I}_1, \hat{I}_2, ..., \hat{I}_N  \} $
\vspace{0.40cm}
\LineComment Find the maximum upper bound of the threshold for which all images are classified as normal
\State $\mathcal{E} = 0 $
\For{ $j = 1$ to $N$  }
\vspace{0.25cm}
\State $\mathcal{E} = Max(\mathcal{E}, \text{MeanSquaredError}(\hat{I}_j, I_j))$
\vspace{0.25cm}
\EndFor
\vspace{0.25cm}
\LineComment Classify the image into normal and abnormal classes
\For{ $\tau = 0$ to $\mathcal{E}$ }
\For{ $j = 1$ to $N$}
    \vspace{0.25cm}
    \State $S(t) = \text{MeanSquaredError}(\hat{I}_j, I_j)$
    \vspace{0.25cm}
    \State $\Phi = 
    \begin{cases}
        Abnormal Frame,& \text{if} S(t) \geq \tau\\
        Normal Frame,& \text{else}
    \end{cases}$
\EndFor
\EndFor 

\end{algorithmic}
\end{algorithm}
The model's output is $\Phi$, the set of normal and abnormal frames. 

    \item \textbf{Receiver Operator Characteristic (ROC)} \\
    There are two criteria for evaluating anomaly detection models: i) The frame level criterion and ii) the pixel level criterion, suggested in prior works \cite{liu2018future,zhou2019anomalynet}. We use two metrics -- i) the Equal Error Rate (EER) \cite{wu2019deep} and ii) the Area under the ROC curve (AUC) \cite{le2022attention}. These two measures are based on the receiver operating characteristics (ROC) curves. We obtain the true positive rate (TPR) and false positive rate (FPR) in the thresholding process. We plot these acquired TPR and FPR values to create the ROC curve. These evaluation metrics are further elaborated in Section \ref{EvalMetrics}.

\end{itemize}

\end{spacing}
\end{Large}



\begin{center}
    \begin{LARGE}
        \textbf{Chapter 3}
        \vspace{-0.5cm}
        \section{Methodology}
        \label{methodology}
    \end{LARGE}
\end{center}

\begin{Large}
\begin{spacing}{1.0}
Our task is to accurately and efficiently predict the occurrences of anomalies in video footage. We propose a novel GAN Architecture that combines some of the latest research work on improving the cognitive abilities of the model. This section provides an in-depth overview of our model, along with vivid visual diagrams to elaborate our research. \\ \\

\subsection{Spatio-Temporal Generative Adversarial Network (STem-GAN)}
The STem-GAN model follows an adversarial autoencoder model. It comprises two main networks, the Generator($\mathcal{G}$) and the Discriminator($\mathcal{D}$). The model is prediction-based and takes the input as a window of video frames. We give a detailed overview of our model below.

\begin{figure}[h]
\centering
\includegraphics[width=0.8\textwidth,height=0.8\textheight,keepaspectratio]{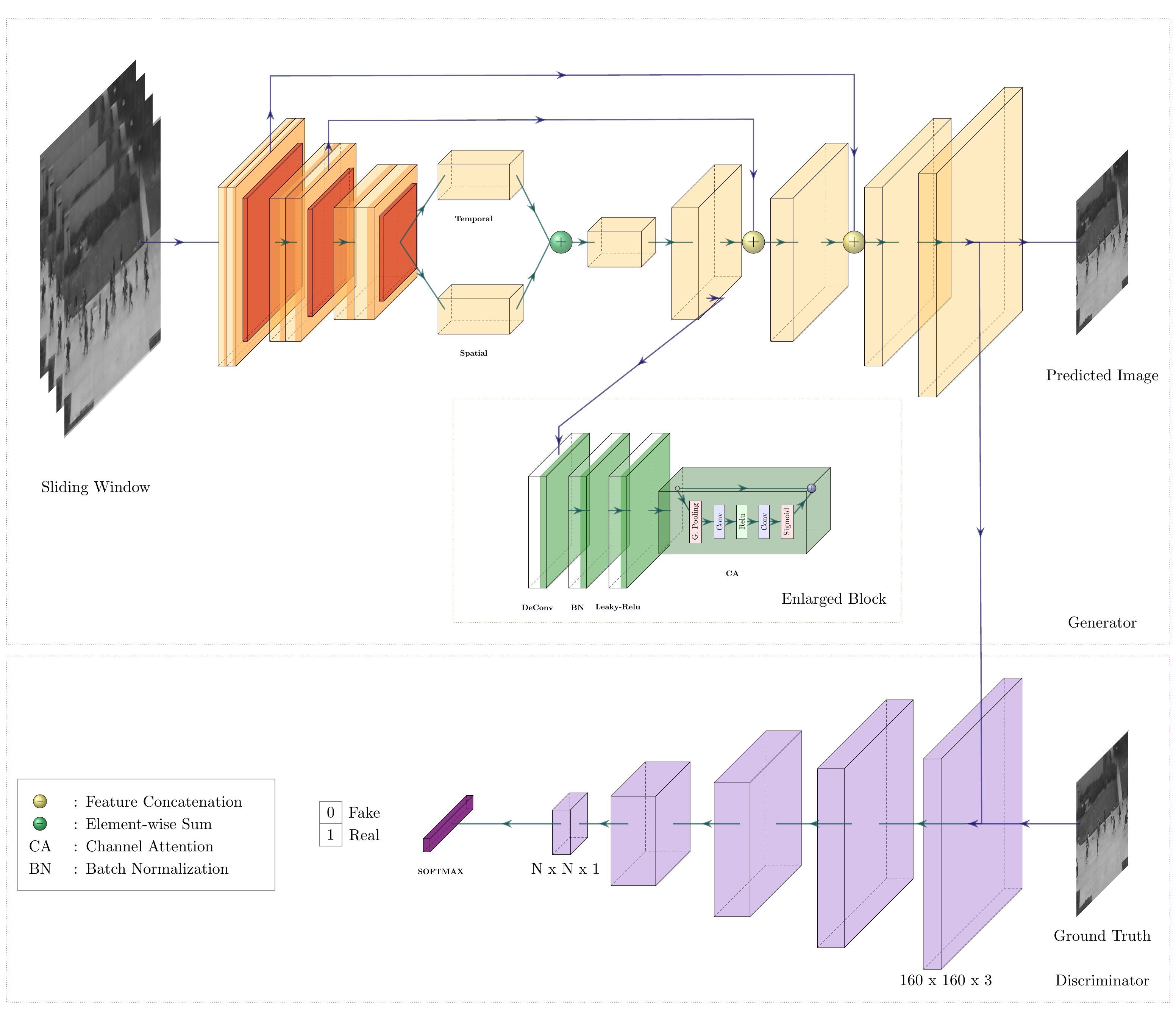}
\captionof{figure}{\large STem-GAN Architecture}\label{Fig:STemGAN}
\end{figure} 

\subsubsection{Generator}
    The generator uses an Autoencoder architecture. It comprises an encoder layer that uses a two-stream channel to extract the spatio-temporal features from the input frames. This layer converts the high-dimensional input to a low-dimensional latent representation. The decoder layer uses the learned parameters and generates the future frame with the input frame's resolution. 
    The layers are explained below:
    
\subsubsection{Encoder}
    The encoder uses a deep convolutional model as its backbone to extract the salient features of the images. This deep learning method is extensively used to learn the pattern and association between images. We experimented with different deep and wide CNN models and chose the \textit{WiderResnet} \cite{wu2019wider} as the suitable backbone. Further, the features extracted from the last layer of deep CNN are split into two streams. This allows our model to extract both the spatial and temporal features of the input images. The temporal branch uses a \textit{temporal shift module} to extract the temporal information from the frames efficiently. In the case of the spatial branch, the inputs are concatenated together to maintain spatial consistency. These features are concatenated and passed to the decoder. These combined features are called -- \textit{Spatio-Temporal Maps} (Figure \ref{Fig:spatio_temporal1}) 

\begin{figure}[h]
\centering
\includegraphics[width=0.5\textwidth,height=0.5\textheight,keepaspectratio]{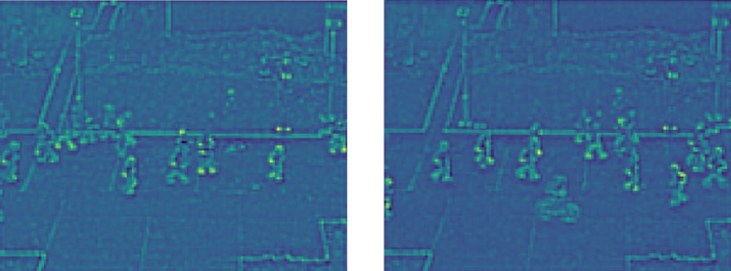}
\captionof{figure}{\large Spatio Temporal Feature Maps}\label{Fig:spatio_temporal1}
\end{figure} 
    
\subsubsection{Decoder}
    The decoder uses deconvolution to upsample the low-dimensional latent space to input resolution. It comprises a $1 \times 1$ convolution layer to reduce the number of parameters. This is followed by a sequence of DeConv blocks. The green region in Figure \ref{Fig:STemGAN} shows the enlarged view. Each DeConv block comprises of stacked Deconvolution layer, Batch Normalization layer, and activation layer in series. Each block is followed by a channel-attention (CA) module as explained in section \ref{sec:attention_pre}. The channel attention module exploits the channel relationships between features and assigns them an implicit weightage, this gives more cognitive power to our model. The channel-attention module was suggested in \cite{hu2018squeeze, woo2018cbam, zhang2018image}. The generated feature maps from this layer are concatenated with encoder features using skip connections. This further helps in stabilizing the training of the generator. 

\subsubsection{Temporal branch}
    The temporal shift method \cite{lin2019tsm} has been applied to video comprehension. In the current work, we want to use the temporal shifting technique to take advantage of temporal information in detecting video anomalies. The temporal dimension is used to accomplish the shift operation. As seen in Fig. \ref{Fig:ts1}, some channels are kept, and some are moved to the following frame.
    
\begin{figure}[h]
\centering
\includegraphics[width=0.50\textwidth,height=0.25\textheight]{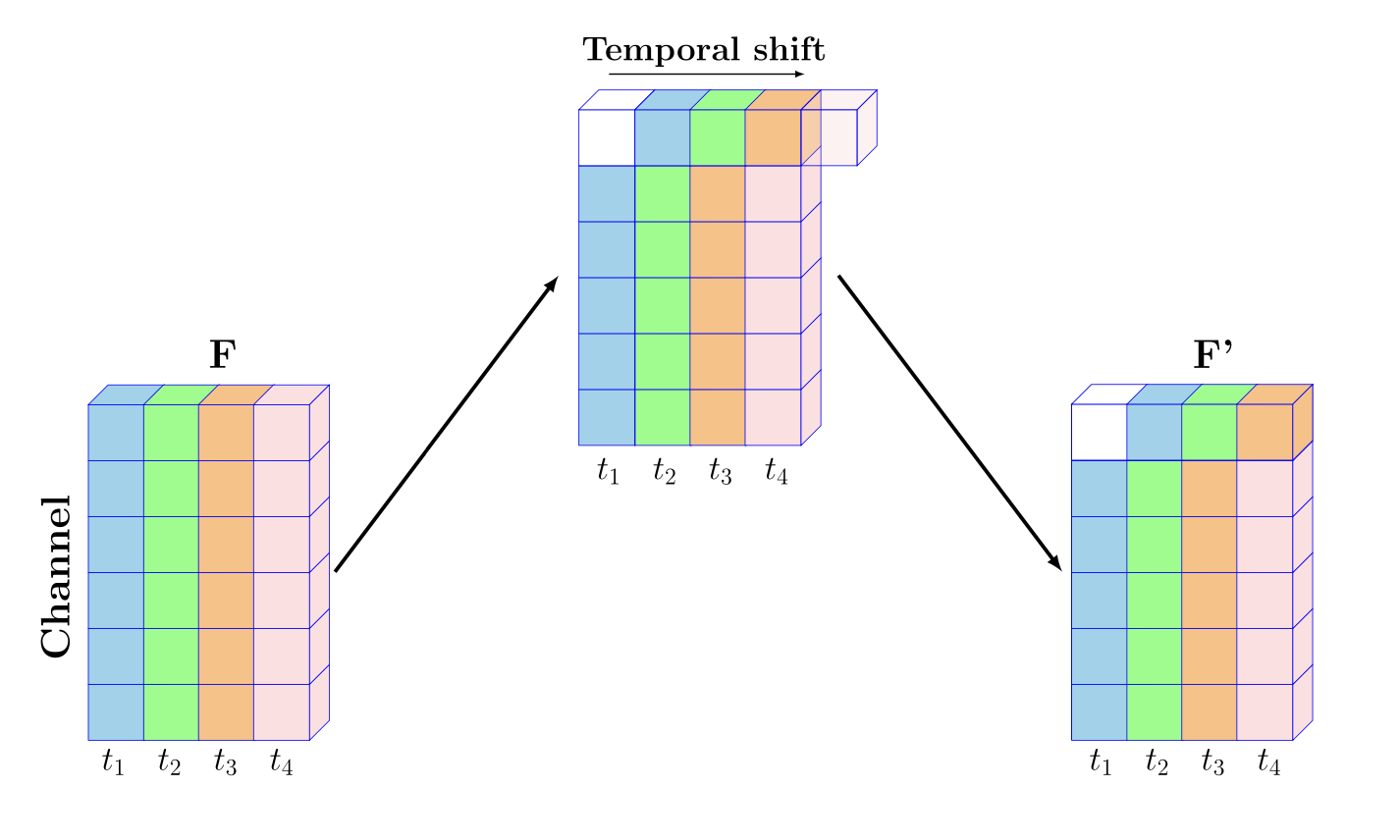}
\captionof{figure}{\large Temporal Shift}\label{Fig:ts1}
\end{figure} 
    Then, the current frame's features and the prior frame's features are combined. The output features are calculated as follows for the given input feature maps
     $\mathbf{F}_{\text {tem }} \in$ $\mathbb{R}^{N \times T \times C \times H \times W}$ are
$$
\mathbf{F}_{\text {tem }}^{\prime}=\operatorname{Shift}\left(\mathbf{F}_{\text {tem }}\right)
$$
where Shift refers to the shift operation. 

\subsubsection{Discriminator}
    The discriminator model functions as the critic for our GAN model. It performs the task of classifying the frames into two classes -- Real and Fake. It learns to distinguish between the ground truth($I_{N+1}$) and the predicted image($\hat{I}_{N+1}$). It bridges the gap between real/fake to normal/abnormal through the reconstruction error concept. It states that the reconstruction error for abnormal images is far greater than for normal ones. Thereby, the discriminator learns to distinguish between the two. \\ \\
    The discriminator follows a convolutional architecture model -- PatchGAN, proposed in Isola et al. \cite{isola2017image}. This patch-based discriminator learns to penalize the structure at a patch level. The model is comprised of back-to-back convolution, batch-normalization and activation layer. The last activation layer is the sigmoid layer. The output is a matrix divided into $N \times N$ patches, each corresponding to a local subregion of the original frame. Isola et al. prove that the model's performance remains fairly consistent even from small values of N. This makes patchGAN a suitable choice for the discriminator.

\subsection{Adversarial Training}
Gan is useful in generation of high-quality reconstruction of images and videos \cite{denton2015deep,mathieu2015deep}. The model consists of a generator $\mathcal{G}$ and a discriminator $\mathcal{D}$. $\mathcal{D}$ learns to distinguish between the ground truth and the predicted frames. While, the $\mathcal{G}$ iteratively learns to generate images that fool the discriminator into predicting these as real. In order to train the two models simultaneously we use the adversarial training. It takes the form of a min-max game between the two models. Ideally, when the generator is well trained, the discriminator cannot classify real and fake better than chance. We fine-tune this training in order to account for a well trained generator that can predict the future frame. The training pipeline is elaborated as follows (Fig. \ref{Fig:train_test}):

\begin{figure}[h]
\centering
\includegraphics[width=0.8\textwidth,height=0.8\textheight,keepaspectratio]{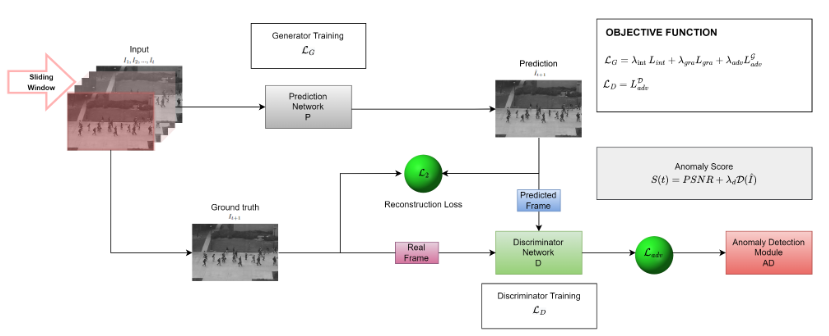}
\captionof{figure}{\large Training-Testing Pipeline}\label{Fig:train_test}
\end{figure} 

\subsubsection{Training Discriminator} The goal of training the discriminator is to learn the binary classification of fake and real images. The Ground truth -- $I_{t+1}$ is classified as $1$, and the generated (fake) image $\hat{I}_{t+1}$ is classified as $0$. We fix the generator weights for training $D$. The binary-cross entropy loss for the real and fake classifications as summed up as shown in Equation \ref{eqn:advD}.
\begin{equation}\label{eqn:advD}
    L_{a d v}^{\mathcal{D}}(\hat{I}, I) =\sum_{i, j}  L_{B C E}\left(\mathcal{D}(I)_{i, j}, 1\right)+\sum_{i, j} L_{B C E}\left(\mathcal{D}(\hat{I})_{i, j}, 0\right)
\end{equation}
Here $i,j$ denotes the 2D index of the patches, and $L_{B C E}$ is the Binary Cross Entropy loss, defined in Equation \ref{eqn:bce}:
\begin{equation}\label{eqn:bce}
L_{B C E}(\hat{X}, X)=-\frac{1}{n} \sum_{i=1}^n\left(X_i \cdot \log \hat{X}_i+\left(1-X_i\right) \cdot \log \left(1-\hat{X}_i\right)\right)
\end{equation}
Here $\hat{X} \in [0,1]$ and $X$ is either 0 or 1.

\subsubsection{Training Generator} The aim of training the generator is to learn the generation of the future frame to such a precision that it will fool $\mathcal{D}$ into classifying it as real. Equation \ref{eqn:advG} shows the adversarial loss for training ${\mathcal{G}}$:

\begin{equation}\label{eqn:advG}
L_{a d v}^{\mathcal{G}}(\hat{I}) =\sum_{i, j}  L_{B C E}\left(\mathcal{D}(\hat{I})_{i, j}, 1\right)
\end{equation}

The convergence of this adversarial loss of $\mathcal{G}$ to 1 is hard to achieve as it essentially means instability in training, and in our experiments, the loss usually converges to $0.50$. Also, $\mathcal{G}$ can always produce samples that can fool $\mathcal{D}$, even if they are not close to the input space $Y$. To address this, we introduce the structural similarity loss, similar to \cite{lu2019future,lu2020few}. We train $\mathcal{G}$ with a combined loss of the adversarial loss (Eqn. \ref{eqn:advG}) and the $L_2$ loss (Eqn.  \ref{eqn:int}).

\begin{equation} \label{eqn:int}
    L_{i n t}(\hat{I}, I)=\|\hat{I} - I\|_2^2
\end{equation}

Therefore, we need to minimize 
$
    \lambda_{\text {int }} L_{i n t}\left(\hat{I}_{t+1}, I_{t+1}\right)+\lambda_{a d v} L_{a d v}^{\mathcal{G}}\left(\hat{I}_{t+1}\right)
$. As per the findings of \cite{mathieu2015deep}, we need to adjust the weights and balance the trade-off between image quality and similarity with ground-truth image samples.

\subsubsection{Image Gradient Constraint Constraints}
In the study of Mathieu et al.\cite{mathieu2015deep}, intensity and gradient differences are used to improve the quality of the predicted image. Our network's objective is to predict the next frame $\hat{I}_{i+1}$, given a sliding window of images $(I_1, I_2,..., I_i)$. Since there are numerous pixels in each frame, most of which have a non-zero intensity, these additional constraints can significantly reduce the prediction error. To be more precise, we reduce the $L_2$ gap in intensity space between a Predicted frame $\hat{I}$ and its truth ground $I$.

A gradient constraint is introduced to the video frame to deal with potential blur when using $L_2$ distance. Following are the steps the loss function takes to determine how different absolute gradients along two spatial dimensions are:

\begin{equation}\label{eqn:gra}
    \begin{aligned}
    L_{g r a}(\hat{I}, I) =& 
    \sum_{i, j} \left\lVert\lvert\hat{I}_{i, j}-\hat{I}_{i-1, j}\rvert-\lvert I_{i, j}-I_{i-1, j}\rvert\right\rVert_1
    + \\&
    \sum_{i, j}\left\lVert\lvert\hat{I}_{i, j}-\hat{I}_{i, j-1}\rvert-\lvert I_{i, j-1}-I_{i-1, j-1} \rvert\right\rVert_1\\
    \end{aligned}
\end{equation}
Here $i$ and $j$ denote the 2D positions of pixels in the image.

\subsubsection{Objective Function}

We take the weighted sum of the constraints on structural similarity and motion with the adversarial constraints to define the objective function for our model. The objective function that we need to minimize is given below.

The objective function for the generator training step:
\begin{equation}
    \begin{aligned}
    L_{\mathcal{G}} &=\lambda_{\text {int }} L_{i n t}\left(\hat{I}_{t+1}, I_{t+1}\right)+\lambda_{g r a} L_{g r a}\left(\hat{I}_{t+1}, I_{t+1}\right)
    +\lambda_{a d v} L_{a d v}^{\mathcal{G}}\left(\hat{I}_{t+1}\right)
    \end{aligned}
\end{equation}

The objective function for the discriminator training step:
\begin{equation}
    L_{\mathcal{D}}=L_{a d v}^{\mathcal{D}}\left(\hat{I}_{t+1}, I_{t+1}\right)
\end{equation}

The parameters $\lambda_{int}$, $\lambda_{gra}$, and $\lambda_{adv}$ are the three coefficients that balance weights between the loss functions.

\subsubsection{Anomaly Score}
To decide normal occurrences accurately, we need a metric that accounts for the prediction capability of our model. As suggested in \cite{liu2018future}, the Mean Squared Error is one such method for evaluating the aggregate pixel-wise quality of predicted images. Peak Signal Noise Ratio (PSNR) is another practical method for evaluating the predicted frame, as demonstrated by Mathieu et al.\cite{mathieu2015deep}, indicated below in Equation \ref{psnr}:

\begin{equation}\label{psnr}
    \operatorname{PSNR}(Y, \hat{Y}) =10 \log _{10} \frac{[\max _{\hat{Y}}]^2}{\frac{1}{N} \sum_{i=0}^N\left(Y_i-\hat{Y}_i\right)^2}
\end{equation}

Here, N is the count of pixels across rows and columns in the image, and ${\max _{\hat{Y}}}$
is the maximum value of 
$\Hat{Y}$

Following \cite{liu2018future}, we normalized the PSNR values of the testing clip to the range [0, 1]. This anomaly score is represented by P(t), as shown in equation \ref{eqn:norm_psnr}:

\begin{equation}\label{eqn:norm_psnr}
P(t)=\frac{P S N R_t-\min (P S N R)}{\max (P S N R)-\min (P S N R)}
\end{equation}

Here, the terms $min(P S N R)$ and $max(P S N R)$ refer to the minimum and maximum PSNR values, respectively. We introduce another anomaly score for STem-GAN, similar to Zenati et al. \cite{zenati2018efficient}. This new score takes a weighted combination of PSNR score and the normalised discriminator scores. Therefore, a lower score would mean a higher probability of an anomaly. It is formulated in Equation \ref{eqn:anomaly_score} below.

\begin{equation}\label{eqn:anomaly_score}
    S(t)= P(t) + \lambda_{d} \mathcal{D}(\hat{I})
\end{equation}
Here P(t) is the normalized PSNR score, and $\lambda_{d}$ is the weight for the discriminator score. The anomaly score S(t) is useful to distinguish between the normal and abnormal frames by statistically determining a threshold for the video. The only requirement is a well-trained discriminator $\mathcal{D}$.

\end{spacing}
\end{Large}
\begin{center}
    \begin{LARGE}
        \textbf{Chapter 4}
        \vspace{-0.5cm}
        \section{Experimental Walkthrough}
        \label{expSetup}
    \end{LARGE}
\end{center}

\begin{Large}
In this chapter, we will explain the datasets used and the experimental setup, along with the specifications of our system used. The section \ref{datadesc} will mention the various datasets used to train and test our proposed method along with a brief overview of its contents.
\begin{spacing}{1.0}
\vspace{0.5cm}
\subsection{Data Description}
\label{datadesc}
We have experimented with various datasets to evaluate the proposed method. We have used the UMN Dataset \cite{bird2006real}, UCSD-Peds Dataset \cite{mahadevan2010anomaly}, CUHK Avenue dataset \cite{lu2013abnormal}, and Subway
Dataset(Entry and Exit) \cite{adam2008robust}. These datasets have been recorded from the CCTVs fixed at a point. The testing frames have frame level annotations as flag bits where 1-bit denotes an anomalous frame and 0-bit a non-anomalous frame.  \\

\textbf{UCSD-Peds:} A fixed camera installed at a height and gazing down on pedestrian pathways was used to collect the UCSD Anomaly Detection Dataset. Bikers, skateboarders, tiny carts, and pedestrians crossing a path or in its surrounding grass are examples of often occurring anomalies. A few incidents involving wheelchair-bound individuals were also noted. All anomalies are real; they weren't produced to create the dataset. Two separate subsets of the data were created, one for each scenario. Each scene's video recording was divided into several segments, each with about 200 frames of size (160,240).  \\
\begin{itemize}
    \item Peds1: Footage showing crowds of people moving in both directions from and toward the camera, with some perspective distortion. It contains 34 examples of training videos and 36 examples of testing videos.
    \item Peds2: Sequences with movement parallel to the camera's axis. 12 testing video samples and 16 training video examples are included.
\end{itemize}
The ground truth annotation for each clip contains a binary flag for each frame, indicating if an abnormality is present in that particular frame. Additionally, manually created pixel-level binary masks are given for a subset of 10 films for Peds1 and 12 clips for Peds2, showing abnormalities' locations. This is done to make it possible to assess how well algorithms perform in terms of their capacity to locate abnormalities. The following images \ref{Fig: Anamoly (Cart) }, \ref{Fig: Anamoly (Biker) } show some anomalies.\\
\begin{figure}[h]
       \centering \includegraphics[width=.60\textwidth,height=.50\textheight,keepaspectratio]{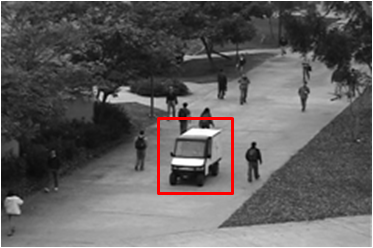}
       \captionof{figure}{\large Anamoly (Cart) \cite{mahadevan2010anomaly}}\label{Fig: Anamoly (Cart) }
  \end{figure}
\begin{figure}[h]
       \centering \includegraphics[width=.60\textwidth,height=.50\textheight,keepaspectratio]{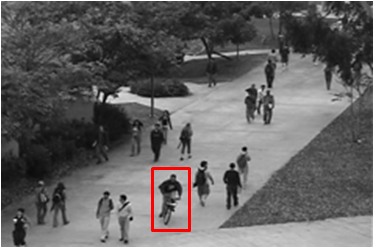}
       \captionof{figure}{\large Anamoly (Biker) \cite{mahadevan2010anomaly}}\label{Fig: Anamoly (Biker) }
  \end{figure}

\textbf{UMN Dataset:} The University of Minnesota's (UMN) crowd dataset is a typical one which includes 11 training scenarios for three different crowd types. Each frame has a 240 x 320 pixel resolution. Each video begins with typical crowd activity and concludes with everyone fleeing the location. The most frequent abnormality in this dataset is the irregular running action due to apprehensions. We can refer to the following image \ref{Fig: UMN dataset } for example.\\
\begin{figure}[h]
       \centering \includegraphics[width=.60\textwidth,height=.50\textheight,keepaspectratio]{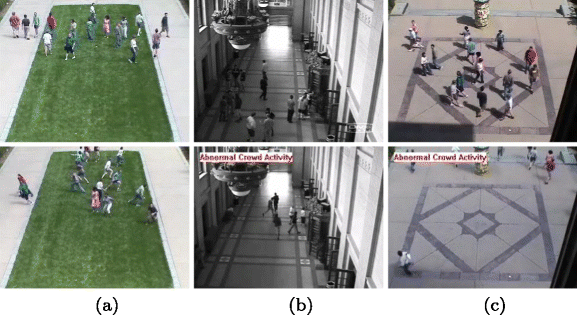}
       \captionof{figure}{\large UMN dataset \cite{bird2006real}}\label{Fig: UMN dataset }
  \end{figure}

\textbf{CUHK Avenue Dataset:} The Avenue dataset has videos in RGB mode. There are a total of 16 training videos comprising 15,328 frames. Then there are also 21 testing videos consisting of 15,324 frames in total. Each frame is 360 × 640 in size. This dataset poses some limitations due to glitches in the camera and shaking. The dataset consists of 47 irregular
events, like running, loitering and walking in opposite directions. We can refer to the following image \ref{Fig: Avenue dataset  } for example.\\
\begin{figure}[h]
       \centering \includegraphics[width=.60\textwidth,height=.50\textheight,keepaspectratio]{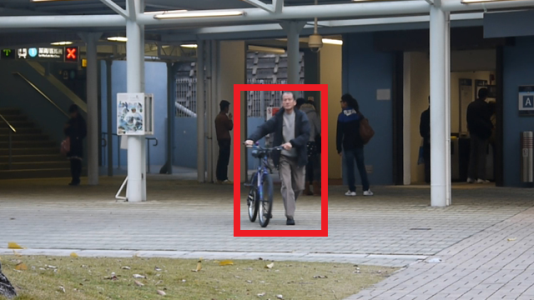}
       \captionof{figure}{\large Avenue dataset \cite{lu2013abnormal}}\label{Fig: Avenue dataset }
  \end{figure}

\textbf{Subway dataset (Entry and Exit):} The Subway dataset has two scenes, the entrance and exit. The Entrance dataset has videos of 1h and 36 min duration accounting for 144249 frames. The Exit dataset has videos of 43 min and 64900 frames. In both datasets, frames are of size 512 × 384 pixels. The Entrance and Exit datasets have 19 types of anomalies, which includes loitering, walking in the wrong directions. Camera glitches, and shaking pose challenges to this dataset. We can refer to the following image \ref{Fig: Subway dataset } for example.\\
\begin{figure}[h]
       \centering \includegraphics[width=.60\textwidth,height=.30\textheight]{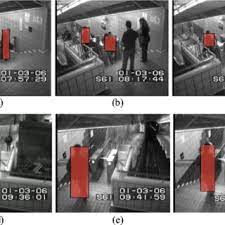}
       \captionof{figure}{\large Subway dataset  \cite{adam2008robust}}\label{Fig: Subway dataset }
  \end{figure}
  
\vspace{0.5cm}

\subsection{Data Pre-Processing} \label{sec:data_preprocessing}

The two major steps in the preprocessing of datasets were to extract frames from the videos using OpenCV and use image processing methods to regularize the images. We built a robust image pipeline to support our framework. It is explained in section \ref{sec:data_preprocessing}.  

\begin{enumerate}
    \item Frame Extraction 
    Most of the datasets are presented as video(AVI) files. To run our anomaly detection module we need to convert these into singular frames in RGB format. We use the OpenCV module video reader to extract the frames using the same fps rate as the original dataset. The fps for datasets are tabulated in Table \ref{tab:dataset_fps}.
    
    \begin{table}[h]
    \begin{center}
    \Large
    \begin{tabular}{|m{8cm}|m{3cm}|m{3.2cm}|} 
    \hline
    \textbf{Dataset} & \textbf{Video Time} & \textbf{FPS} \\
    \hline
    UMN & 4 mins & 25 \\
    \hline
    UCSDpeds & 10 mins & 10 \\
    \hline
    Avenue & 50 mins & 15 \\
    \hline
    Subway (Entrance \& Exit) & 120 mins & 20 \\
    \hline
    \end{tabular}
    \end{center}
    \caption{\large FPS rate of datasets}
    \label{tab:dataset_fps}
    \end{table}

    \item Label Extraction
    We performed a comprehensive analysis of the video frames and developed the ground truth labels pertaining to videos. For the testing videos, the abnormal frames have been labelled as 1, and the normal frames are labelled 0: meaning that there is no anomaly. The labels are confirmed with the actual labels (if present). We also provide the critically examined, standardized labels for anomaly events in unlabelled datasets (UMN and Subway dataset).
\end{enumerate}

\subsection{Data Pipeline}
We need to represent videos as a set of frames. We have used OpenCV for frame extraction. Then each frame is resized to 160 x 160 resolution to satisfy the requirements of our model. To improve the performance and training stability of our model, we introduced a normalization layer in our pipeline that shifts and scales inputs into a 0-centred distribution with a variance of 1. Normalization also ensures that each input parameter (pixel, in this case) is bounded to $[-1, 1]$. This makes convergence faster while training the network. We also created methods for extracting a 3D window of frames with customizable strides and dividing the image into 3D-cuboidal patches of size $20 \times 20$. Figure \ref{Fig:pipeline} better explains the pipeline in a visual manner.\\ \\

\begin{figure}[h]
   \centering \includegraphics[width=0.2\textwidth,height=0.4\textheight]{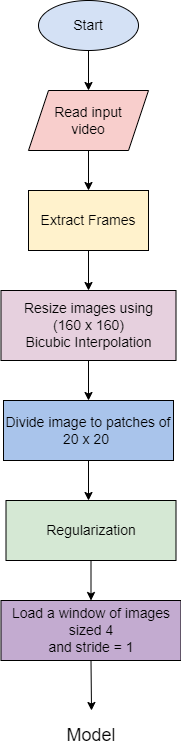}
   \captionof{figure}{\large Pipeline }\label{Fig:pipeline}
\end{figure}

\vspace{1cm}
Further, We have optimized the I/O pipeline by introducing generic IO optimization techniques. We have created a buffer to load only a section of frames into the RAM, this reduces the load on the memory by asynchronously loading the image from the disk into RAM. Additionally, We use prefetching and parallelizing methods to take advantage of the temporal localization of video frames. We note down the ablation study in Table \ref{tab:IOperformance}.

\begin{table}[h]
    \begin{center}
    \Large
    \begin{tabular}{|m{3cm}|m{2cm}|m{2cm}|m{2cm}|m{2cm}|} 
    \hline
    Caching & $\checkmark$  & $\checkmark$ & -- & $\checkmark$  \\
    \hline
    Prefetching & -- & $\checkmark$ & $\checkmark$ & $\checkmark$ \\
    \hline
    Parallelizing & -- & -- & $\checkmark$ & $\checkmark$ \\
    \hline
    FPS & 8.5 & 10.1 & 12.5 & 15.4\\
    \hline
    \end{tabular}
    \end{center}
    \caption{\large IO speedup}
    \label{tab:IOperformance}
    \end{table}

\subsection{Dataset Analysis}

The first stage of our progress was to analyse the distribution of our dataset. First, we used the \textbf{Yolo-v5} based object detection model pretrained on a subset of \textbf{Coco} dataset to identify the distinct objects and develop the bounding boxes over the identified objects. In this stage, we parse the scene and shortlist the hypothetical foreign objects not infused with the scene background. This allows us to perform a background agnostic analysis capable of subsiding some of the complexity of multi-scene AD.
We sample normal objects from the training dataset and develop a non-parametric object database, as depicted below.

\begin{figure}[h]
\centering
\includegraphics[width=0.6\textwidth,height=0.6\textheight,keepaspectratio]{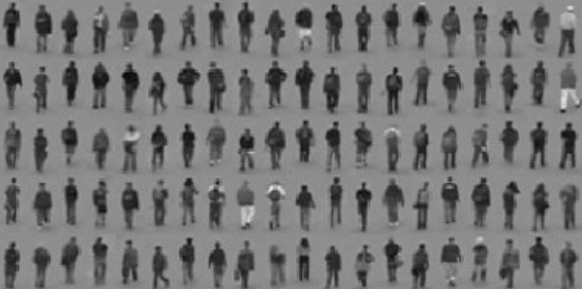}
  \captionof{figure}{\large Non-parametric dataset of usual objects }
  \label{Fig:normal_yolo} 
\end{figure}

\begin{figure}[h]
\centering
\includegraphics[width=0.8\textwidth,height=0.8\textheight,keepaspectratio]{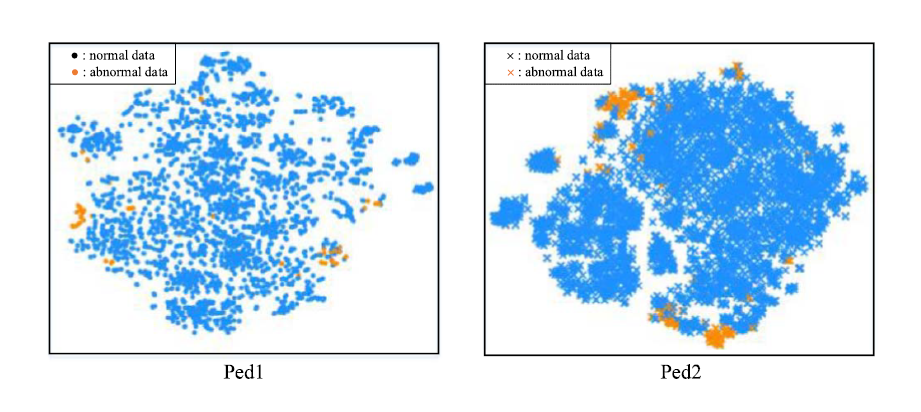}
\captionof{figure}{\large T-SNE scatter plot for Ped1 \& Ped2 dataset}\label{Fig:features2}
\end{figure} 

\noindent We extract the features from our trained STem-GAN model for the Peds1 and Peds2 dataset. We plotted a scatterplot of our results in Figure \ref{Fig:features2}.


\subsection{Hardware and Software Specifications}
Our experiments were carried out on the local server with the following specifications. Intel Xeon processor loaded with Linux Operating System. The machine harboured 128GB of RAM, along with an Nvidia Tesla V100 GPU graphics card with 16GB of Memory and 15.7 TFLOPS in deep learning performance. We also tested our model on Nvidia P100 and T4 GPU-managed notebooks provided by Google Cloud. As noted in our experiments, our model size is over 6GB, and together with the data pipeline, it consumes 7GB of GPU memory. \\ \\

The complete code is written and tested in the Pytorch framework. The datasets were extracted and stored in two folders: the training folder, containing only normal images and the testing folder, containing the test videos. The models were saved and exported as default \textit{pth} files.

Chapter \ref{expResults} will explain the results, and the various 
hyperparameters decided upon.
\end{spacing}
\end{Large}
\begin{center}
    \begin{LARGE}
        \textbf{Chapter 5}
        \vspace{-0.5cm}
        \section{Experimentation and Results}
        \label{expResults}
    \end{LARGE}
\end{center}

\begin{Large}
\begin{spacing}{1.0}
In this section, we will test our model against other contemporary models in video anomaly detection. We take a quantitative approach towards our comparative study. We give suitable reasoning for the evaluation metrics we have chosen. We provide a detailed overview of our experimentation setup -- the distinct constants and learning parameters we have used to get our results.

\vspace{0.5cm}
\subsection{Performance Evaluation Metrics}
    \label{EvalMetrics}
We have used a frame-level criterion to determine the anomaly detection capability of our model. Our metrics include the Area under the ROC curve (AUROC) score and the Equal error rate (EER). These two measures are generic classification metrics for a model, unlike accuracy, precision etc. which are dependent on a particular threshold. We first plot the receiver operating characteristic (ROC) curve using the TPR and FPR values at different thresholds.

\begin{equation}
    \text{TPR} = \frac{\text{Number of true positives}}{\text{Number of total positives}}
\end{equation}

\begin{equation}
    \text{TPR} = \frac{\text{Number of false negatives}}{\text{Number of total negatives}}
\end{equation}

\begin{enumerate}
    \item \textbf{AUROC Score} \\
    It is a useful metric to evaluate the ability of a classifier to separate between classes. We can visualize it as the area under the ROC curves. Usually, a higher score, closer to $1.0$, is considered a good model, whereas a score close to $0.5$ means that the model is guessing randomly. In another case, a score towards $0.0$ signifies that the model is predicting the exact opposite of the results. These cases are shown in ROC plots below.

    \item \textbf{EER score} \\
    This metric provides a complete summarization of the classification ability of the model. It deals with the ratio of wrongly classified frames in the model. In the context of frame-level criterion, this corresponds to the point where $TPR = 1 - FPR$. We can extract this score through interpolation methods. Generally, a lower EER is preferred because it signifies higher accuracy.
\end{enumerate}

\subsection{Experimental Setup}
We used a method similar to \cite{le2022attention} to train our model. We adopted the Adam optimizer with a learning rate of $2e-4$, for both $\mathcal{G}$ and $\mathcal{D}$. The hyperparameters $\beta1$ and $\beta2$ were set to $0.5$ and $0.999$ respectively. The Adam optimizer is an improvement of the normal SGD optimizer, it provides better results for computer vision tasks. We have set the anomaly score parameter $\lambda_d$ to 0.3 to get the best results. To properly train both 
$\mathcal{G}$ and $\mathcal{D}$, we use a technique suggested by \cite{pfau2016connecting}. We train the discriminator five times for each generator iteration. The training continues till the discriminator score converges to $0.5$ and $ ||\hat{I} - I||^2_2 < 0.001$. The model is trained on a maximum of 60 epochs for any dataset. The input frames are passed as window frames of the size of 5 and stride of 1.
\subsection{Quantitative Evaluation}
\label{sec:quantitative}
The model is tested on the five benchmark datasets. The results we obtained are compared to the state-of-the-art models. The results are compiled below in the form of a comparison study. We note that our model outperforms the state-of-the-art models by a significant margin, proving the efficiency and capability of our model in real-time anomaly detection tasks.


The model is trained using the same experimentation technique for all datasets. The tables \ref{table:auc_umn}, \ref{table:auc_peds}, and \ref{table:auc_subway} show that our model outperforms the others in UCSDped2, UMN and the Subway Exit dataset. While for the other datasets -- CUHK Avenue and the Subway Entrance dataset, we observe a reasonable score, which is slightly reduced due to the increasing complexity of the datasets. We analyse the trends in our obtained results below.

\begin{enumerate}
    \item \textbf{Complexity of dataset} \\
    Our results show a direct relationship between the complexity of the anomaly with the AUROC score. For datasets with a simpler set of anomalies like running, walking in the wrong direction and entrance of cycles etc. For example, the UMN, UCSDped2 and the subway exit dataset. Whereas in scenarios with more subtle anomalies like loitering, jumping etc. There are more chances of a wrong prediction. This challenge is faced in the avenue and subway entrance datasets. Another issue is the camera angles and resolution of the image. This challenge is faced in the UCSDped1 dataset.

    \item \textbf{Perspective and point of interest} \\
    In any video surveillance footage, we can split the video into multiple regions -- the active regions and the passive setup environment. The active region is usually a small video section where we normally find functional events. On the other hand, the passive regions refer to the stationary, unchanging elements, such as buildings, trees etc. The perspective refers to the angle of the trajectory of the objects with the camera. A parallel pathway is more suitable than a skewed one to detect anomalies. For example, we achieve a $97.5\%$ AUROC score for Ped2 (parallel walkway) than $81.2\%$ for Ped1 (Skewed walkway).
\end{enumerate} 


\begingroup
\setlength{\tabcolsep}{10pt}
\renewcommand{\arraystretch}{1.0}
\begin{table}[h]
\Large
\begin{center}
\begin{tabular}{|m{9.0cm}|m{3.0cm}|}
\hline
Model & AUROC (\%)\\
\hline 
Social force (Mehran et al.)\cite{mehran2009abnormal} & 96.0\\
Sparse (Cong et al.) \cite{cong2011sparse}
& 97.5\\
Local aggregate (Saligrama and Chen)\cite{saligrama2012video} & 98.5\\
Chaotic invariants (Wu et al.)\cite{wu2010chaotic} & 99.0\\

Ravanbakhsh et al.(2017) \cite{ravanbakhsh2017abnormal} & 99.0\\
Ours Scene1 & \textbf{99.74}\\
Ours Scene2 & \textbf{99.20}\\
Ours Scene3 & \textbf{99.70}\\
\hline
\end{tabular}
\end{center}
\caption{\large Comparison of the AUROC score for UMN dataset.}
\label{table:auc_umn}
\end{table}
\endgroup

\begingroup
\setlength{\tabcolsep}{10pt}
\renewcommand{\arraystretch}{1.2}
\begin{table}[H]
\Large
\begin{center}
\begin{tabular}{|m{5.0cm}|m{2cm}|m{2cm}|m{2cm}|m{2cm}|  }
\hline
 & \multicolumn{2}{c|}{UCSDped2}  & \multicolumn{2}{c|}{CUHK Avenue}\\
 \cline{2-5}
 \centering{ Methods} & AUROC & EER & AUROC & EER\\

\hline
Tang et al. \cite{tang2020integrating} & 96.3 & 10.0 & 85.1 & --\\

 Wei et al. \cite{wei2019detecting} & 89.5& 14.2 & 79.7 & 23.0\\
 Nawaratne et al. \cite{nawaratne2019spatiotemporal} & 91.1& 8.9 & 76.8 & 29.2\\
 Zhou et al. \cite{zhou2019attention} & 96.0& -- & 85.78 & --\\
 Yang et al. \cite{yang2020improving} & 95.9& 11.1 & 85.9 & 20.4\\
 Chang et al. \cite{chang2020clustering} & 96.5 & -- & 86.0 & --\\
 Abati et al. \cite{abati2019latent} & 95.4 & -- &– & --\\
 Fang et al. \cite{fang2020multi} & 95.6 & -- & 86.3 & --\\
 Gong et al. \cite{gong2019memorizing} & 94.1 & -- & 83.3 & --\\
 Luo et al. \cite{luo2019video} & 92.2& -- & 83.5 & --\\
 Liu et al. \cite{liu2018future} & 95.4 & -- & 85.1 & -- \\
 Zhou et al. \cite{zhou2019anomalynet} & 94.9& 10.3 & 86.1 & 22.0\\
 Lu et al. \cite{lu2019future} & 96.0& -- & 85.7 & --\\
 Li et al. \cite{li2019spatio} & 96.5 & 8.7 & 84.5 & 21.5\\
 Deepak et al. \cite{deepak2021residual} & 83.0& -- & 82.0 & --\\
 Doshi et al. \cite{doshi2021online} & 97.2& -- & 86.4 & --\\
  Le et al. \cite{le2022attention} & 97.4 & -- & \textbf{86.7} & --\\
 Ravanbakhsh et al. \cite{ravanbakhsh2019training} & 95.5& 11.0 & -- & --\\
 STem-GAN(Ours) & \textbf{97.5} & \textbf{6.23}& \textbf{86.0} & \textbf{20.1}\\

\hline
\end{tabular}
\end{center}
\caption{\large Comparative analysis of the AUROC score and EER for UCSDpeds2 and Avenue}\label{table:auc_peds}
\end{table}
\endgroup

\begingroup
\setlength{\tabcolsep}{10pt}
\renewcommand{\arraystretch}{1.0}
\begin{table}[h]
\large
\begin{center}
\begin{tabular}{|m{6.0cm}|m{2.3cm}|m{2.3cm}|}
\hline
Method & Entrance & Exit\\
\hline  
MDT14 \cite{li2013anomaly} & 89.7 & 90.8\\
SRC \cite{cong2011sparse} & 80.2 & 83.3\\
Conv-AE \cite{hasan2016learning} & \textbf{94.3} & 80.7\\
LSTM-AE \cite{luo2017remembering} & 93.3 & 87.7\\
Unmasking \cite{tudor2017unmasking} & 71.3 & 86.3\\
NMC \cite{ionescu2018detecting} & 93.5 & 95.1\\
DeepOC \cite{wu2019deep} & 91.1 & 89.5\\
Luo et al. \cite{luo2019video} & 85.4 & 89.7\\
STem-GAN(Ours) & \textbf{90.4} & \textbf{95.2}\\
\hline
\end{tabular}
\caption{\large Comparative analysis of the AUROC score for Subway dataset.}\label{table:auc_subway}
\end{center}
\end{table}
\endgroup


\subsection{Visualization}

\begin{figure}[h]
    \centering
    \includegraphics[width=1.0\textwidth,height=0.4\textheight]{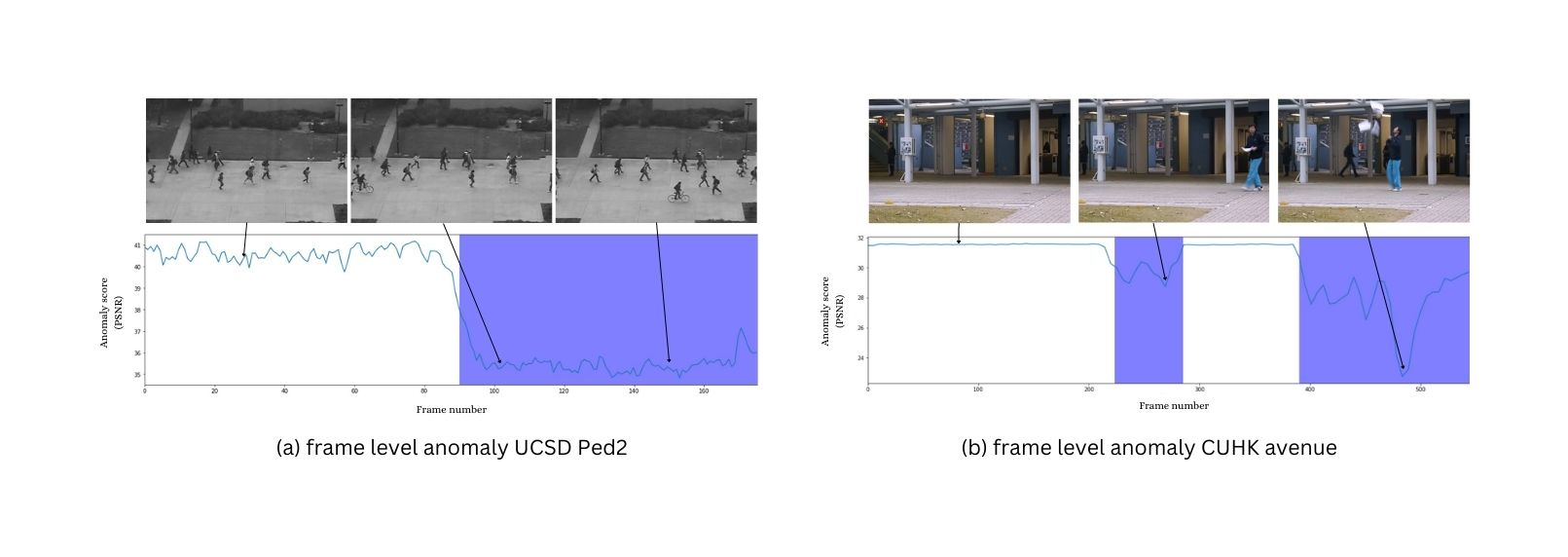}
    \captionof{figure}{\large Frame Anomaly Scores}\label{Fig:frame_anomaly_score} 
  \end{figure}

Figure \ref{Fig:frame_anomaly_score} shows the plot of the anomaly score (S(t)) across the video. We follow the blue line, which tracks the anomaly score. The purple region signifies the labelled region of anomalies in the video. We also notice that in some regions, there are substantial changes in the slope of the line. On further investigation, one can identify the cause of this change is the appearance of abnormal events. \\
In Figure \ref{Fig:frame_anomaly_score}(a), we display the graph for UCSDpeds2 (testing video 02). We observe that the blue line drops abruptly starting from the second image. A bicycle has entered the frame in this region, marked as purple. \\
Further, in Figure \ref{Fig:frame_anomaly_score}(b), we show the anomaly scores for CUHK Avenue (testing video 13). This video shows a staged anomaly where a person first enters the scene from the wrong direction and then loiters the area. There are two regions marked as purple, which means two abnormal events take place. Our model can generalize and handle both of these cases with high precision. 
\vspace{0.5cm}
\subsection{Impact of Different Losses}
We perform the study over distinct choices of loss functions for training the Stem-GAN model. The model was tested over various combinations of structural, gradient and adversarial loss. The results are tabulated in Figure \ref{Fig:losses_ablation}. We compare the results using the AUROC score and a new metric, the Score Gap, as proposed in Liu et al. \cite{liu2018future}.  The score gap is represented using the symbol $\Delta_s$, and it denotes the gap between the mean anomaly score of Normal vs Abnormal frames. It is formally formulated below.

\begin{equation}\label{eqn:score_gap}
    \Delta_s = \frac{1}{n}\sum_{j=1}^{n} \text{Normal-S(t)}_j
    - \frac{1}{m}\sum_{j=1}^{m} \text{Abnormal-S(t)}_j
\end{equation}

Where $n$ and $m$ are the counts of regular and irregular frames, respectively, and $S(t)$ represents the anomaly score as formulated in Equation \ref{eqn:anomaly_score}.
The greater the $\Delta_s$, the better our model is able to distinguish between normal and abnormal frames. The results from the table prove that with more constraints, our model can perform better.

\begin{figure}[h]
    \centering
 \includegraphics[width=1.0\textwidth,height=0.6\textheight,keepaspectratio]{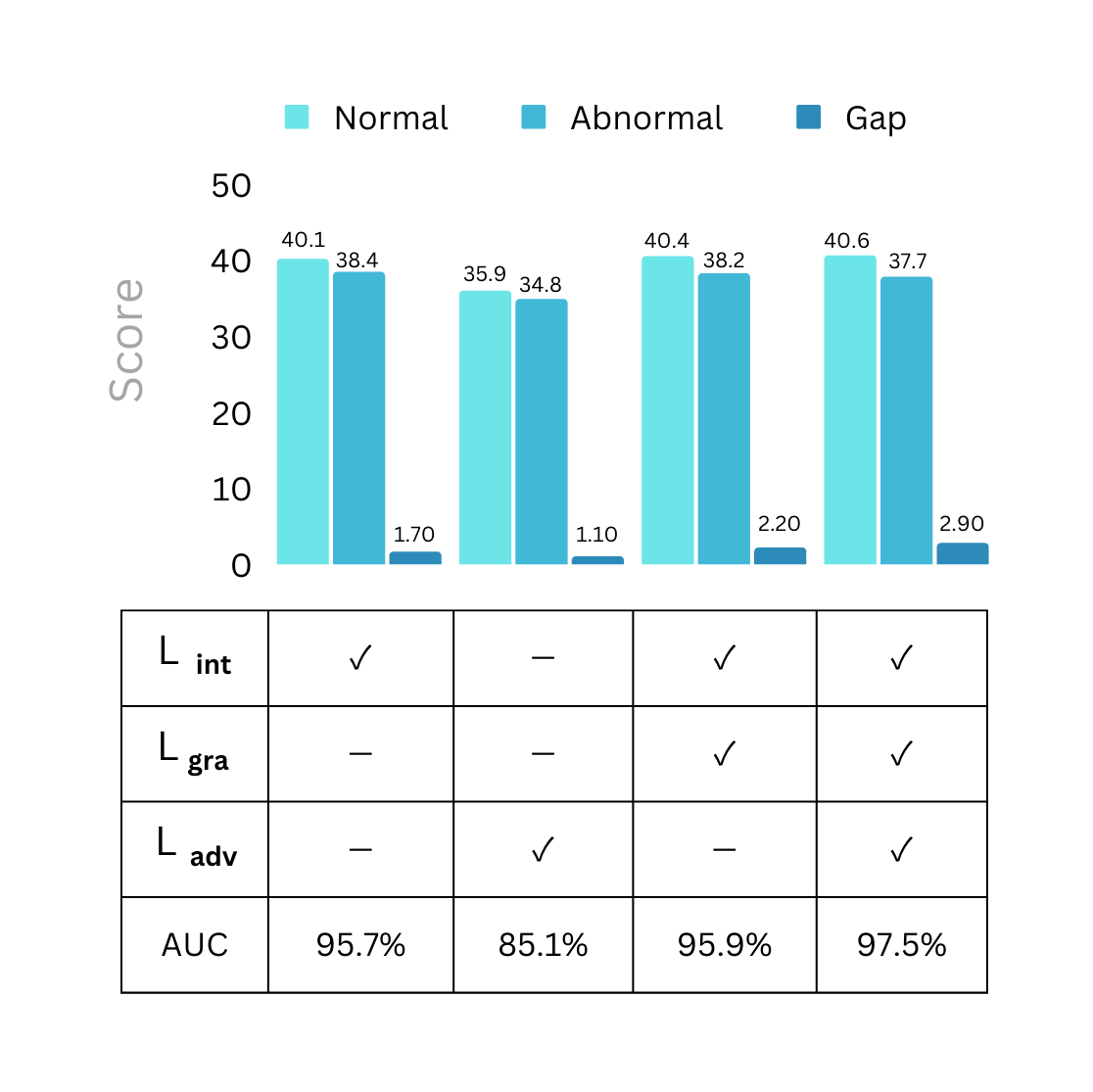}
    \captionof{figure}{\large Comparison of Loss Functions}\label{Fig:losses_ablation} 
  \end{figure}

\subsection{Transfer Learning}
In this section, we display how our model, transfer learning of our model, can improve the generalization and training time of the anomaly detection model. We initialize the optimizers with a learning rate of 1e-4, half of that taken before, keeping the other parameters the same. Table \ref{table:transfer} shows how the model trained on one dataset proves to be useful for the other. Here $A \rightarrow B$ means training of $B$ wrt weights of $A$. The model gives better results in a lower number of epochs in most cases. We observe that the new model takes a lower number of iterations to reach the same level of quality. In addition, our model learns to identify new patterns and features in the same dataset. Thus we prove that transfer learning is an efficient way to train models on datasets with similar characteristics, like colour schemes and scene conditions.

\begin{table}[h]
\Large
\begin{center}
\begin{tabular}{|m{8.0cm}|m{3cm}|m{3cm}|}
\hline
Dataset & Non-Transfer learning & Transfer learning\\
\hline
$\text{UMN Scene3} \rightarrow \text{UMN Scene1}$ & \textbf{99.74\%} & 99.70\% \\
$\text{UMN Scene1} \rightarrow \text{UMN Scene2}$ & 99.20\% & \textbf{99.57\%} \\
$\text{UMN Scene1} \rightarrow \text{UMN Scene3}$ & 99.70\% & \textbf{99.72\%} \\
$\text{UCSDped2} \rightarrow \text{UCSDped1}$ & 79.80\%  & \textbf{81.20\%} \\
$\text{UCSDped1} \rightarrow \text{UCSDped2}$ & 97.50\% & \textbf{97.54\%}\\
\hline
\end{tabular}
\end{center}
\caption{\large GAN-based transfer learning}
\label{table:transfer}
\end{table}

\subsection{Analysis of StemGAN}

In this section, we provide a practical analysis of STem-GAN on benchmark datasets. This study allows us to get a qualitative insight into our model.

\begin{enumerate}
    \item \textbf{Analysis of Frame-Rate} \\
    We report the average runtime of various contemporary models along with their machine configurations. From table \ref{table:fps}, It is easy to see that our model ranks in the middle of the list. Our model does not utilize any additional model to calculate the optical flow, giving it an advantage over such models. Considering our model's deep and wide CNN layers, it successfully provides a reasonable frame rate of 11 fps.

\begingroup
\setlength{\tabcolsep}{10pt}
\renewcommand{\arraystretch}{1.2}
\begin{table}[h]
\Large
\begin{center}
\begin{tabular}{|m{4cm}|m{3cm}|m{5cm}|m{2cm}|}
\hline
Method & Platform & GPU & FPS\\
\hline  
MDT14 \cite{li2013anomaly} & C & -- & 0.90\\
ADMN \cite{xu2017detecting} & Matlab & Nvidia Quadro k40000 & 0.10\\
SRC \cite{cong2011sparse} & Matlab & -- & 0.22\\
Liu \cite{liu2018future} & Tensorflow & Nvidia Titan & 20\\
Unmasking \cite{tudor2017unmasking} & -- & -- & 20\\
NMC \cite{ionescu2018detecting} & -- & -- & 23.8\\
DeepOC \cite{wu2019deep} & Tensorflow & Nvidia P100 & 40\\
STem-GAN & \textbf{Pytorch} & \textbf{Nvidia V100} & \textbf{11}\\
\hline
\end{tabular}
\end{center}
\caption{\large Comparison of the Frame-Rate(fps)}\label{tbl:fps}
\label{table:fps}
\end{table}
\endgroup

    \item \textbf{Quality of Predicted Images} \\
    Although our model provides state-of-the-art results for performance metrics like AUROC and EER scores, we provide more evidence for the practicality of our model. In Figure \ref{Fig:reconstruction_error}, we have shown the ground truth and predicted image along with its Mean Squared Error. Although our model has potentially blurry edges due to inherently misformed predictions, the MSE score is much lower than other CNN frameworks \cite{wu2019deep}.
\end{enumerate}

\begin{figure}[h]
    \centering
 \includegraphics[width=0.6\textwidth,height=0.4\textheight]{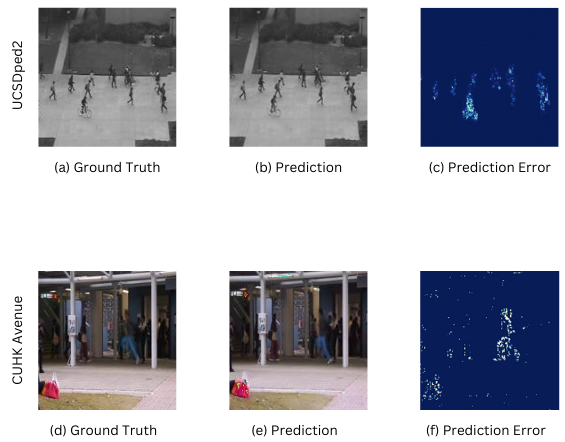}
    \captionof{figure}{\large Reconstruction Error}\label{Fig:reconstruction_error} 
  \end{figure}

\end{spacing}
\end{Large}
\begin{center}
    \begin{LARGE}
        \textbf{Chapter 6}
        \vspace{-0.5cm}
        \section{Conclusion and Future Scope}
    \end{LARGE}
\end{center}

\begin{Large}
\begin{spacing}{1.00}
\vspace{0.5cm}
\subsection{Conclusion}
This work has given us a better understanding of the steps needed to create a machine learning model, including the analysis phase, understanding the need for real-world use cases, and the in-depth technical knowledge needed to develop the machine learning model.
We have used the novel Spatio-temporal GAN model developed for detecting anomalies in various datasets. This model also uses attention for predicting future frames in a given video. This is a self-sufficient model because of its implicit temporal shift method used to learn the information of motion. To guarantee that the predicted image goes in hand with the ground truth in the case of regular events, we have included additional structural similarity constraints. Images with higher anomaly scores can thus be categorised as anomalous. Extensive testing has revealed that our model outperforms current state-of-the-art anomaly-detecting models, demonstrating the power of our approach. Our model performs well for AUROC scores and other useful measures, according to the ablation research.

\subsection{Future Scope}
Specific types of abnormal events have been tested using the methods presented in this thesis. More varieties of anomalous occurrences can be tested using various large datasets. By expanding the scope of our research, we intend to use a real-world highway dataset to train our model and identify anomalous situations like exceeding the posted speed limit, driving on the wrong side of the road, and persons crossing the road unlawfully. Emotional traits might also be considered when classifying abnormal events for future research because changes in people’s emotions typically come before abnormal occurrences. 
\end{spacing}
\end{Large}
\newpage\thispagestyle{empty}
    \mbox{}
    \newpage
\bibliographystyle{plain}

{\large
\bibliography{main.bib}}
\end{sloppypar}
\end{document}